%% file: main.tex
\documentclass{article}

\usepackage{microtype}
\usepackage{graphicx}
\usepackage{subfig}
\usepackage{booktabs} % for professional tables
\usepackage[hyphens]{url}

\usepackage{hyperref}

\newcommand{\ie}{\textit{i.e.}}
\newcommand{\eg}{\textit{e.g.}}

\newcommand{\methodname}{\textbf{PIMMUR}}
\newcommand{\numpaper}{\color{myblue}350}
\newcommand{\numstudy}{\color{myblue}576}
\newcommand{\numfail}{\color{myblue}91.1\%}
\newcommand{\numperfect}{\color{myblue}14, 2.4\%}
\newcommand{\avgU}{\color{myblue}65.2\%}
\newcommand{\avgM}{\color{myblue}50.6\%}

\usepackage[T1]{fontenc}
\usepackage{colortbl}
\usepackage{arydshln}
\usepackage{enumitem}
\usepackage{multirow}
\usepackage[subfigure]{tocloft}
\usepackage[most]{tcolorbox}
\usepackage{pifont}
\usepackage{upquote}
\newcommand{\cmark}{\textcolor{mygreen}{\ding{51}}}
\newcommand{\xmark}{\textcolor{myred}{\ding{55}}}

\definecolor{myred}{RGB}{139, 0, 0}
\definecolor{mygreen}{RGB}{0, 100, 0}
\definecolor{myblue}{RGB}{0, 0, 0}
\definecolor{myyellow}{RGB}{253, 253, 143}

\tcbset{
  aibox/.style={
    top=8pt,
    bottom=3pt,
    colback=white,
    colframe=black,
    colbacktitle=black,
    enhanced,
    breakable,
    center,
    attach boxed title to top left={yshift=-0.1in,xshift=0.15in},
    boxed title style={boxrule=0pt,colframe=white,},
  },
}
\newtcolorbox{AIbox}[3][]{aibox, width=#2, title=#3,#1}

\usepackage{soul}
\sethlcolor{myyellow}
\newif\ifrevshow \revshowfalse
\newcommand{\rev}[1]{\ifrevshow\hl{\textbf{#1}}\else{\color{myblue}#1}\fi}
\newcounter{revcounter}
\newcommand{\DefineRevision}[3]{
  \stepcounter{revcounter}
  \expandafter\edef\csname revnum#1\endcsname{\arabic{revcounter}}
  \expandafter\def\csname revorig#1\endcsname{#2}
  \expandafter\def\csname revnew#1\endcsname{#3}}
\newcommand{\revnumof}[1]{
  \ifcsname revnum#1\endcsname\csname revnum#1\endcsname\else\textbf{??}\fi}

\newcommand{\revDelBody}[1]{\begingroup\sethlcolor{myblue}\color{myblue}\hl{#1}\endgroup}
\newcommand{\InsertRevision}[1]{\protect\hypertarget{rev:#1}{}
  \expandafter\ifx\csname revnew#1\endcsname\empty
    \expandafter\expandafter\expandafter\revDelBody
    \expandafter\expandafter\expandafter{\csname revorig#1\endcsname}
  \else
    \expandafter\ifx\csname revorig#1\endcsname\empty
      {\color{myblue}\csname revnew#1\endcsname}
    \else
      \csname revnew#1\endcsname
    \fi
  \fi}

\newcommand{\revfield}[3]{
  \expandafter\ifx\csname rev#1#2\endcsname\empty #3\else\csname rev#1#2\endcsname\fi}

\input{Responses/revisions}
\usepackage[preprint]{icml}

\usepackage{amsmath}
\usepackage{amssymb}
\usepackage{mathtools}
\usepackage{amsthm}

\usepackage[capitalize,noabbrev]{cleveref}

\theoremstyle{plain}

\theoremstyle{definition}

\theoremstyle{remark}

\usepackage[textsize=tiny]{todonotes}

\icmltitlerunning{The {\methodname} Principles}

\begin{document}

\clearpage
\onecolumn

\addtolength{\cftsecnumwidth}{1.5em}
\addtolength{\cftsubsecnumwidth}{1.5em}
\setlength{\cftsubsecindent}{3em}
\addtocontents{toc}{\protect\setcounter{tocdepth}{3}}
\setcounter{page}{1}
\setcounter{section}{0}
\renewcommand{\thesection}{\Roman{section}}

\clearpage
\twocolumn

\addtocontents{toc}{\protect\setcounter{tocdepth}{0}}
\setcounter{page}{1}
\setcounter{section}{0}
\renewcommand{\thesection}{\arabic{section}}

\twocolumn[
    \icmltitle{The {\methodname} Principles: \\ Ensuring Validity in Collective Behavior of LLM Societies}
    % \icmltitle{The Illusion of Emergent Social Behavior: \\ A Systematic Audit of Methodological Flaws in AI Society Simulations}
    % \icmltitle{A Systematic Audit of LLM-Based Social Simulations Finds Unsupported Claims about Human Collective Behavior}
    
    % the [accepted] option to the icml package.
    
    
    \icmlsetsymbol{equal}{*}
    
    \begin{icmlauthorlist}
        \icmlauthor{Jiaxu Zhou}{equal,cuhk}
        \icmlauthor{Jen-tse Huang}{equal,jhu} \\
        \icmlauthor{Xuhui Zhou}{cmu}
        \icmlauthor{Man Ho Lam}{cuhk}
        \icmlauthor{Xintao Wang}{fdu}
        \icmlauthor{Hao Zhu}{stanford}
        \icmlauthor{Wenxuan Wang}{ruc,rai}
        \icmlauthor{Maarten Sap}{cmu}
    \end{icmlauthorlist}
    
    \icmlaffiliation{cuhk}{CUHK}
    \icmlaffiliation{jhu}{JHU}
    \icmlaffiliation{cmu}{CMU}
    \icmlaffiliation{fdu}{Fudan}
    \icmlaffiliation{stanford}{Stanford}
    \icmlaffiliation{ruc}{RUC}
    \icmlaffiliation{rai}{renlab.ai}
    
    \icmlcorrespondingauthor{Wenxuan Wang}{jwxwang@gmail.com}
    \icmlcorrespondingauthor{Maarten Sap}{maartensap@cmu.edu}
    
    \icmlkeywords{Machine Learning, ICML}
    
    \vskip 0.3in
]



\printAffiliationsAndNotice{\icmlEqualContribution}

\begin{abstract}
\input{Sections/0_Abstract}
\end{abstract}

\input{Sections/1_Introduction}
\input{Sections/2_Results}
\input{Sections/3_Discussion}
\input{Sections/4_Methods}

\input{Sections/5_Statements}

\bibliography{model,reference,mass}
\bibliographystyle{icml}

%%%%%%%%%%%%%%%%%%%%%%%%%%%%%%%%%%%%%%%%%%%%%%%%%%%%%%%%%%%%%%%%%%%%%%%%%%%%%%%
%%%%%%%%%%%%%%%%%%%%%%%%%%%%%%%%%%%%%%%%%%%%%%%%%%%%%%%%%%%%%%%%%%%%%%%%%%%%%%%
% APPENDIX
%%%%%%%%%%%%%%%%%%%%%%%%%%%%%%%%%%%%%%%%%%%%%%%%%%%%%%%%%%%%%%%%%%%%%%%%%%%%%%%
%%%%%%%%%%%%%%%%%%%%%%%%%%%%%%%%%%%%%%%%%%%%%%%%%%%%%%%%%%%%%%%%%%%%%%%%%%%%%%%

\clearpage
\onecolumn
\appendix
\input{Sections/Appendix}

%%%%%%%%%%%%%%%%%%%%%%%%%%%%%%%%%%%%%%%%%%%%%%%%%%%%%%%%%%%%%%%%%%%%%%%%%%%%%%%
%%%%%%%%%%%%%%%%%%%%%%%%%%%%%%%%%%%%%%%%%%%%%%%%%%%%%%%%%%%%%%%%%%%%%%%%%%%%%%%

\end{document}

%% file: Responses/revisions.tex
\DefineRevision{intro-define-scope}
{The central research question is to examine whether patterns observed in human societies can also emerge in societies composed purely of LLM agents.
These simulations promise a transformative in \textit{silico} laboratory for computational social science, offering a scalable testbed for sociological theories that are otherwise difficult or unethical to test in the real world.}
{Such simulations are motivated by four broad objectives (Fig.~\ref{fig:scope}): \rev{testing whether human social behaviors emerge spontaneously in LLMs, a property of the AI itself (upper right); using LLMs as proxies for humans to test whether a given factor elicits a particular behavior, a property of human behavior (upper left); engineering AI societies to exhibit a target behavior through prompting, where spontaneous emergence is immaterial (lower left); and characterizing the behavior of AI societies of any kind (lower right).
We focus on the first two.}}

\DefineRevision{intro-current-drawbacks}
{}
{Moreover, existing works often rely on customized frameworks tailored to specific social experiments~\cite{yang2024oasis, borah2025mind}, and methodological choices vary widely and lack shared standards~\cite{zhou2024real}.
It is therefore unclear whether their reported findings reflect genuine emergent social phenomena, or artifacts of flawed experimental design.}

\DefineRevision{intro-inductive-deductive}
{\rev{Specifically, we} identify six recurring pitfalls that span the micro-design of agents and the macro-design of the environment, which we formalize as the {\methodname} framework:}
{\rev{To investigate this problem, we first conducted a pilot review of 39 papers and} identified six recurring pitfalls that span the micro-design of agents and the macro-design of the experimental environment, which we formalized as the {\methodname} principles:}

\DefineRevision{intro-systematic-review}
{In this study, we conduct a systematic audit of 39 high-profile LLM social simulation papers published between 2023 and 2025.
We find that the current ``AI society'' literature suffers from a pervasive crisis of methodological rigor.
Existing works often rely on customized frameworks tailored to specific social experiments~\cite{yang2024oasis, borah2025mind}, but methodological choices vary widely and lack shared standards~\cite{zhou2024real}.
As a result, it is unclear whether their reported findings reflect genuine emergent social phenomena, or artifacts of flawed experimental design.}
{\rev{Guided by the principles above, we conducted a PRISMA-compliant systematic audit of {\numstudy} MASS studies reported across {\numpaper} papers published between December 2022 and May 2026.
We found a pervasive mismatch between the employed methods and the emergent AI behaviors they are claimed to demonstrate.}}

\DefineRevision{results-inductive-deductive}
{To systematically evaluate the state of LLM-based MASS studies, we propose the {\methodname} framework.
These six principles---\textit{Profile, Interaction, Memory, Minimal-Control, Unawareness}, and \textit{Realism}---are derived from a synthesis of methodological shortcomings identified across our audit of {\numpaper} recent studies (\S\ref{sec:review}).}
{To systematically evaluate current LLM-based MASS studies, we propose the {\methodname} principles.
These six---\textit{Profile, Interaction, Memory, Minimal-Control, Unawareness}, and \textit{Realism}---are derived from a \rev{pilot review of 39 MASS papers, as detailed in} \S\ref{sec:pilot}.}

\DefineRevision{results-profile-mincontrol}
{Note that if the \textit{Profile} is intentionally biased to induce certain outcomes (\eg, making most agents highly aggressive in a study of general populations), this would indeed violate \textit{Minimal-Control}.}
{\rev{Whether adherence to \textit{Profile} violates \textit{Minimal-Control} depends on the research goal.
Assigning highly selfish dispositions to most agents would violate \textit{Minimal-Control} in a general study of behavior in the public goods game, but would be compliant if the research question is precisely how selfish individuals behave in that setting.
Similarly, providing an anchor in an anchoring-bias study is not a violation, whereas adding instructions such as ``pay attention to this item'' alongside the anchor is.}}

\DefineRevision{results-systematic-audit}
{To empirically assess the prevalence of the {\methodname} violations, we conducted a systematic audit of representative studies.
We restricted our scope to research employing multi-agent systems to simulate collective human behavior, excluding applications focused purely on task-oriented benchmarks (\eg, software engineering or math problem solving).
After further excluding whose prompts are not available, such as \citet{orlando2025can, tang2025gensim, zheng2024simulating, he2023homophily}, we included 39 papers, where a remarkable 89.7\% violated at least one {\methodname} principle.
Table~\ref{tab:review} summarizes the results, including their simulation goals, selected LLMs, and compliance with the principles.
The criteria for selecting papers and assessing compliance are provided in \S\ref{sec:selection} and \S\ref{sec:extraction}, respectively.
\textbf{PIM} principles are more frequently satisfied than \textbf{MUR}.
Only four papers~\cite{zhang2025trendsim, yang2024oasis, liu2025mosaic, park2023generative} achieve full compliance.
These ``outliers'' typically involve large-scale environments with expansive action spaces that effectively mask the underlying experimental intent.
This widespread non-compliance suggests that the current AI Society literature may suffer from systemic internal validity crises, where reported ``social emergence'' is indistinguishable from instruction-following.}
{To assess the prevalence of {\methodname} violations empirically, \rev{we conducted a systematic audit following the PRISMA 2020 guidelines}~\cite{page2021prisma}.
\rev{Searches of four databases (Scopus, IEEE Xplore, ACM Digital Library, and arXiv) yielded 2,575 records after deduplication.
Screening identified {\numstudy} studies, drawn from {\numpaper} papers, for inclusion in this review.}
Fig.~\ref{fig:flow-diagram} summarizes the identification, screening and inclusion process.
Details of the search and screening procedure are given in \S\ref{sec:selection}, \rev{and the data-extraction protocol and the rules mapping extracted data onto compliance or violation labels in} \S\ref{sec:extraction}.
\rev{We classified the simulation goals of the {\numstudy} studies into five categories comprising 28 domains; descriptions and representative studies for each domain are provided in} \S\ref{sec:taxonomy}.
Table~\ref{tab:review} \rev{reports the number of studies per domain together with the compliance rate for each principle.
Because some papers omitted information that could not be extracted, compliance rates were computed as \textsc{true}/(\textsc{true}+\textsc{false}) and therefore rest on different values of $N$.
\textbf{PIM} principles were satisfied more often than \textbf{MUR} principles.
Notably, {\numfail} of papers were explicitly labeled \textsc{false} on at least one principle, whereas only} {\numperfect} \rev{satisfied all six.
This widespread non-compliance indicates that many claims in the current MASS literature are not supported by the experimental designs used to justify them, making the ``emergent behavior'' an illusion.}}

\DefineRevision{results-unawareness}
{Table~\ref{tab:unawareness-result} shows the results using five frontier models: GPT-4o-2024-11-20, Gemini-2.5-Flash-05-20, Claude-4.0-Sonnet, LLaMA-4-Scout, and Qwen3-235B-A22B.
Stronger models such as GPT-4o correctly identified the simulation’s social objective in 56.4\% of cases.
Notably, LLMs frequently outperformed human annotators in recognizing classic experimental paradigms due to their extensive pre-training on social science literature.
This indicates a ``Curse of Knowledge''~\cite{li2025curse}: the models are often ``too informed'' to serve as unbiased social proxies, as their responses are likely confounded by their internal representation of the expected experimental outcome.}
{Table~\ref{tab:unawareness-result} shows the results using six frontier models: GPT-5.6-Terra~\cite{gpt56sol-gpt56terra-gpt56luna}, Claude-Sonnet-5~\cite{claudes5}, Gemini-3.7-Flash~\cite{gemini37flash}, DeepSeek-V4-Pro~\cite{deepseekv4pro-deepseekv4flash}, Qwen3.8-2.4T-A95B~\cite{qwen38}, and Kimi-K3~\cite{kimik3}.
\rev{Across the six LLMs, the simulation goal was correctly recognized in {\avgU} of cases.
This figure differs from the one reported in} Table~\ref{tab:review}, \rev{which requires a binary judgment for each study: there, we record a violation of \textit{Unawareness} when at least half of the LLMs recognize the simulation goal.
Compliance differs markedly across the five categories.
C1 shows the lowest compliance rate (15.8\%), as it consists mainly of classical game-theoretic scenarios that LLMs recognize readily.
C2 (opinion and information dynamics) follows, while C3 and C4---both concerning interpersonal interaction---are comparable.
C5 shows the highest compliance, but this reflects its composition rather than better practice: it comprises mainly psychological and cognitive assessments rather than social simulations.
Notably, our pilot review of 39 papers assessed \textit{Unawareness} using earlier model versions (\eg, GPT-4o, Gemini-2.5, Claude-4), yielding an average recognition rate of 50.8\%.
The increase with newer models indicates that the problem intensifies as model capability grows: more capable models are less suitable for replicating these experiments.}
This pattern is consistent with a ``Curse of Knowledge''~\cite{li2025curse}---models are often too well informed to serve as unbiased social proxies, as their responses are likely confounded by their internal representation of the expected experimental outcome.}

\DefineRevision{results-mincontrol}
{}
{Agreement with the human verdict using 20 randomly sampled papers (33 simulations), measured by Cohen's $\kappa$, was 0.61 for Kimi-K3, 0.61 for Qwen3.8, 0.54 for DeepSeek-V4, 0.47 for Claude-5, 0.46 for GPT-5.6, and 0.24 for Gemini-3.7 (mean 0.49).}

\DefineRevision{results-exp-awareness-citation}
{}
{Such experimental awareness is also documented by \citet{barrie2025emergent}, who found that LLMs frequently recognized coordination games as experimental tasks.
An analogous form of leakage arises with human participants: \citet{orne1959nature} reported that the effect of interest emerged only among subjects who could articulate the experimenter's hypothesis, and \citet{carter1991economists} found that economics students were more likely to reach equilibrium outcomes in coordination games.}

\DefineRevision{results-claim-check}
{}
{These papers span a range of simulation goals, and all have released complete code to reproduce their results.
Each violates either Minimal-Control or Unawareness, and each makes claims, in its abstract and discussion, about whether the observed LLM social behavior closely approximates human behavior in real-world settings.
These papers were selected for the reasons above, rather than for their influence or citation counts.}

\DefineRevision{results-telephone-game-correction}
{Intriguingly, the trajectory of the {\methodname}-compliant group more closely mirrored the Reverse-Control baseline (no statistical difference; $\beta = 0.044$; $z = 1.755$; $P = 0.079$; $n = 100$) than the Original steered group in several rounds.}
{\rev{There is no statistical difference between} the trajectory of the {\methodname}-compliant group and the Reverse-Control baseline ($\beta = 0.044$; $z = 1.755$; $P = 0.079$; $n = 100$).
\rev{Moreover, Ours shows narrower error bars than both Original and Reverse, consistent with the absence of steering: instructing LLMs to be either accurate or inaccurate reduces the diversity of transmitted messages.}}

\DefineRevision{results-herd-effect-correction}
{Agents are presented with a multiple-choice question and an aggregated peer choices (\eg, ``You noticed $k$ agents chose A, $l$ agents chose D'').}
{Agents are presented with a multiple-choice question and \rev{an aggregated statement of peer choices} (\eg, ``You noticed $k$ agents chose A, $l$ agents chose D'').}

\DefineRevision{results-social-network-growth-interaction}
{In natural social systems, individuals rarely have global, real-time access to the exact degree of all potential peers.}
{In natural social systems, individuals rarely \rev{encounter peers reduced to a name and a degree.
Beyond online platforms, where follower counts are explicitly displayed, degree is typically not directly observable at all.}}

\DefineRevision{results-social-network-growth-correction}
{Without manual intervention, our simulation achieved a power-law exponent closer to the empirical value of $-2$ (observed in Twitter datasets), with a significantly higher goodness-of-fit ($R^2 = 0.93$) compared to the original design's $R^2 = 0.56$ (Fig. \ref{fig:social-network}).
Formal interaction analysis ($Y \sim X + \text{Group} + X \times \text{Group}$) confirmed that the redesign fundamentally altered the network growth trajectory, revealing a significant interaction effect ($\beta_{\text{int}} = 0.0017$; $p = 0.019$; $n = 300$) and a substantial shift in the baseline formation rate ($\beta_{\text{group}} = -0.0798$; $p = 0.011$; $n = 300$).
This suggests that when agents are granted communicative agency and shielded from experimental intent, the resulting network growth reflects authentic social behavior rather than mere adherence to prompted heuristics.}
{As shown in Fig.~\ref{fig:social-network}, \rev{both designs are compatible with a power-law tail that cannot be rejected (Original: $\alpha = 1.898$, $D_\text{KS} = 0.101$, $P_\text{boot} = 0.644$; Ours: $\alpha = 2.132$, $D_\text{KS} = 0.089$, $P_\text{boot} = 0.824$), and the two exponents are statistically indistinguishable ($\Delta\alpha = 0.234$; 95\% CI $[-0.37, 0.84]$).}
Within the design space explored by \citet{de2023emergence}, \rev{a broad degree distribution emerged under only two conditions: when agents observed peers' degrees directly, at the cost of requiring renaming, or when degrees were withheld, leaving names as the sole available information.
Our redesign recovers a broad degree distribution in a more realistic setting, in which agents acquire information by conversing with one another rather than through information scarcity or name shuffling.}}

\DefineRevision{results-prompt-sensitivity}
{To ensure that the observed behavioral shifts are attributable to {\methodname} violations rather than idiosyncratic prompt sensitivity~\cite{ye2026stop}, we performed a semantic-preserving perturbation test on the Belief Convergence task (\S\ref{sec:fake-news}).
We introduced lexical changes (five-word modifications) to the ``Original,'' ``Ours,'' and ``Reverse'' prompts to match the magnitude of lexical variance required to implement our principles.
As illustrated in Fig. \ref{fig:fake-news-skeptical-rephrase};\ref{fig:fake-news-infected-rephrase};\ref{fig:fake-news-recovered-rephrase}, while these perturbations induced minor baseline shifts---such as a higher global tendency toward skepticism---the structural findings remained invariant.
The discrepancy in the number of infected agents persisted even in the rephrased versions, with a statistically significant difference observed between the original and our simulations (mean difference $= 16.1$; 95\% CI: $[8.2, 24.1]$; $t(17) = 4.48$; $P = 0.009$).
Crucially: (1) the performance gap between the ``Original'' and ``Ours'' conditions persisted across all paraphrased versions; and (2) behaviors in the ``Ours'' condition consistently aligned more closely with the ``Reverse'' control than with the ``Original'' setting.
This suggests that the reported social phenomena in prior studies are not robust to methodological rigor and are likely artifacts of prompt-driven biases.}
{To ensure that the observed behavioral shifts are attributable to {\methodname} violations rather than idiosyncratic prompt sensitivity~\cite{ye2026stop}, we performed a semantic-preserving perturbation test on \rev{the fake news propagation and the social network growth tasks.
We introduced comparable lexical changes to the prompts to match the magnitude of lexical variance required to implement our principles.}

For fake news propagation, while these perturbations induced minor baseline shifts---such as a higher global tendency toward skepticism---the structural findings remained invariant (Fig. \ref{fig:fake-news-skeptical-rephrase};\ref{fig:fake-news-infected-rephrase};\ref{fig:fake-news-recovered-rephrase}).
The discrepancy in the number of infected agents persisted even in the rephrased versions, with a statistically significant difference observed between the original and our simulations (mean difference $= 16.1$; 95\% CI: $[8.2, 24.1]$; $t(17) = 4.48$; $P = 0.009$).
Crucially: (1) the performance gap between the ``Original'' and ``Ours'' conditions persisted across all paraphrased versions; and (2) behaviors in the ``Ours'' condition consistently aligned more closely with the ``Reverse'' control than with the ``Original'' setting.
This suggests that the reported social phenomena in prior studies are not robust to methodological rigor and are likely artifacts of prompt-driven biases.

\rev{For social network growth, the power-law tail remains non-rejectable in both conditions ($N = 300$): the original design gives $\alpha = 2.142$, $D_\text{KS} = 0.069$, $P_\text{boot} = 0.965$, and our redesign gives $\alpha = 2.233$, $D_\text{KS} = 0.097$, $P_\text{boot} = 0.581$.
Both bootstrap $P$ values remain well above the conventional threshold of 0.10, as in the unperturbed run, and the two exponents remain close to one another} (Fig. \ref{fig:social-network-rephrase}).
The conclusion is therefore not an artifact of the particular wording we used.}

\DefineRevision{results-qualitative-analysis}
{}
{Consider social balance as a first example.
When the experimental intent was transparent, agents frequently displayed a ``Silicon Hawthorne Effect,'' explicitly invoking Heider's social balance theory in their internal reasoning when deciding how to relate to others.
Under our framework, by contrast, LLM agents were markedly more affiliative and sought to preserve existing ties: an agent befriending two mutual enemies typically remarked on the imbalance, yet maintained both friendships rather than taking a side, thereby departing from Heider's rule.
Fake news propagation offers a second example.
When prompted that ``humans tend to exhibit confirmation bias,'' agents reproduced this instruction verbatim in their reasoning and endorsed the fake news, rather than reasoning from their assigned role or drawing on their own knowledge.
Once the cue was removed, agents frequently exhibited \textbf{epistemic vigilance}, citing the need for ``independent verification'' and ``cross-referencing'' before accepting a claim, and were more likely to invoke knowledge acquired during pretraining.
This skepticism was most pronounced for scientific misinformation (\eg, equatorial hurricanes): under \textit{Minimal-Control}, agents reasoned from elementary geographic knowledge to reject the claim, and nearly all simulations reached universal disbelief within five rounds.}

\DefineRevision{discussion-lesson-learned}
{}
{A central challenge in replicating these experiments is keeping the LLM agents unaware of the simulation's goal.
The original implementations typically query the agent directly about the construct of interest.
For example, ``<describing relationships to other agents>... Will your new opinion of Agent 1 be negative or positive?''
Such direct questioning, together with the binary relationship label, makes the social-balance setting readily identifiable to the model.
In such cases, we should avoid direct questions (\eg, ``write about your opinion toward other agents''), imposing no binary constraint and giving no cue to the underlying social theory.}

\DefineRevision{discussion-scope}
{We do not propose {\methodname} as a rigid, universal checklist for all computational modeling.
We recognize a fundamental trade-off between theoretical simplicity and ecological validity.
Many pioneering works deliberately simplify agent memory or interaction protocols to isolate specific causal mechanisms.
Such ``minimalist'' simulations remain profoundly valuable for mechanistic understanding and hypothesis generation.
However, {\methodname} becomes a necessary normative standard when researchers move beyond abstract modeling to make human-centric claims---specifically, assertions that AI societies replicate, predict, or yield direct insights into real-world human collective dynamics.
We distinguish \textit{Realism} from the other five structural dimensions. While the \textbf{PIMMU} dimensions ensure the internal validity of the simulation environment, \textit{Realism} provides the external anchor.
In domains where empirical human data are scarce, \textit{Realism} should serve as a guiding aspiration for validation.
Findings derived from simulations that violate these principles must be framed strictly as ``model-specific behaviors'' rather than ``universal social laws.''}
{\rev{MASS is a broad field that encompasses diverse research objectives.
Here we highlight two representative classes of studies.
(1) Studies of social behavior in which particular human factors are explicitly programmed into the agents, as is common in mechanism studies (\eg, evacuation dynamics during disasters, or collective decision-making within a specified population).
Such work falls within our scope, and the programmed factors do not violate \textit{Minimal-Control}, because specifying them is itself part of the research objective.
(2) Studies that ask whether a given machine collective exhibits a particular behavior, where the organization of the collective need not mirror any real-world human counterpart.
This class also covers traditional agent-based modeling (\eg, Sugarscape).
Our principles do not apply to such work.
Finally, the principles address complementary questions: PIM concern whether the system faithfully replicates the real-world situation of interest; MU concern whether the observed behavior is externally imposed or genuinely emerges internally; and R specifies the referent against which claims are made (\eg, human society).}}

\DefineRevision{discussion-limitation}
{}
{The first limitation concerns reproduction.
It is infeasible to reproduce every study in order to verify the effect of enforcing {\methodname} compliance.
Relatedly, the most accurate way to determine \textit{Minimal-Control} compliance would be to test whether removing a given sentence renders the experiment unrunnable; across a review of {\numstudy} studies, assessing compliance from the textual prompts alone is therefore the only feasible approach.
Moreover, our {\methodname}-compliant framework requires interaction, which in some cases increases experimental cost; this is also why we adopted $N = 300$ for the social network growth reproduction rather than the original $N = 2{,}500$.
Our implementation is likewise not the only {\methodname}-compliant one: we operationalize interaction as direct messaging, whereas the definition also admits indirect interaction, such as modification of the shared environment.
Second, although we show that LLMs are susceptible to experimenter bias, we do not propose a means of suppressing model knowledge to mitigate this effect.
One direction is to use LLMs trained solely on historical corpora, such as Talkie-13B~\cite{levine2026talkie-13b}, Ranke-4B~\cite{gottlich2025ranke-4b}, or TypewriterLM-7B~\cite{luo2026typewriterlm-7b}, which should not encode knowledge of social experiments devised subsequently (\eg, after 1950).
Behavior emerging from such models would reflect intrinsic tendencies rather than recall of the experimental design.}

\DefineRevision{method-overview}
{}
{To map the current landscape of LLM-based MASS, we first developed {\methodname} on the basis of a pilot review of 39 papers retrieved from Semantic Scholar.
We then conducted a multi-stage systematic review following the Preferred Reporting Items for Systematic Reviews and Meta-Analyses (PRISMA) statement~\cite{page2021prisma}.
Finally, we replicated five representative social simulations using a {\methodname}-compliant simulation framework.}

\DefineRevision{method-pilot-review}
{}
{To develop the principles, we conducted a pilot review comprising a multi-stage literature search and screening procedure in Semantic Scholar.
A query combining the keywords ``LLM,'' ``collective behavior,'' ``emergence,'' and ``simulation'' returned $n = 944$ candidate records.
Title screening excluded non-LLM research, review articles and position papers, retaining $n = 123$ records.
Abstract screening further excluded studies of task-oriented benchmarks (\eg, software engineering or mathematical problem solving) and recommender systems that did not use simulation to investigate social phenomena, retaining $n = 63$ records.
Full-text review then retained only studies that explicitly compared LLM simulations with human social dynamics and reported reproducible prompts, yielding a final corpus of $n = 39$ papers.
Two authors drafted the {\methodname} principles, and another two authors verified that no additional drawbacks were present across the 39 papers.
In total, 35 of 39 papers (89.7\%) violated at least one principle.
Table~\ref{tab:pilot-review} reports {\methodname} compliance for each paper.}

\DefineRevision{method-registration}
{}
{Our systematic review was pre-registered at OSF on 28 Jun 2026 (available via the following URL: \url{https://osf.io/nf7mv/overview?view_only=c74ea84e0b3f434395681cbb42e8d144}).
The registration was modified once before the screening stage, with five changes specified below and in the document.
We have also provided the original registration document in the supplementary information.}

\DefineRevision{method-registration-modification}
{}
{\begin{enumerate}[noitemsep]

    \item Semantic Scholar was removed as an information source, and arXiv was added. The final set of sources is Scopus, IEEE Xplore, ACM Digital Library, and arXiv.
    \textbf{Reason}: We assumed that Semantic Scholar could be queried using our pre-specified Boolean search strategy. On implementation, we found that it provides no batch retrieval API supporting Boolean operators, so the registered strategy could not be applied to this source. The available alternatives---issuing manual keyword-by-keyword queries or approximating the registered string---would have yielded a record set that was neither reproducible nor systematically comparable with the other sources. We therefore removed the source rather than applying a materially different search to it. Because a substantial share of relevant work in computer science appears first as preprints, removing Semantic Scholar would have reduced our coverage of recent and non-indexed work; arXiv was added to retain that coverage. arXiv supports field-based Boolean querying, so the registered strategy could be applied to it directly.
    \textbf{Impact}: The substitution alters the composition of the retrieved set rather than only its size, reducing coverage of the cross-disciplinary venues indexed by Semantic Scholar and increasing coverage of preprints. Preprints were screened against the same eligibility criteria as all other records.

    \item The search string was revised and expanded relative to the version provided at registration.
    \textbf{Reason}: Pilot searches indicated that the registered string failed to retrieve known relevant papers.
    \textbf{Impact}: The revised string retrieves a broader set of records. All retrieved records were screened against the eligibility criteria, which remained unchanged.

    \item Zotero was used as the reference management tool; no such tool was specified at registration.
    \textbf{Reason}: The volume of retrieved records exceeded expectations and required systematic management. Zotero was used to import records from all four sources, remove duplicates, and attach full texts.
    \textbf{Impact}: Zotero was used for record management only; it played no role in applying the eligibility criteria.

    \item Within the profile-related data items, the open-ended field \texttt{other\_profile\_information} was removed and replaced by three Boolean fields:
    \begin{enumerate}[noitemsep]
        \item \texttt{has\_background\_story},
        \item \texttt{has\_personality}, and
        \item \texttt{has\_role}.
    \end{enumerate}
    The three registered Boolean fields---
    \begin{enumerate}[noitemsep]
        \item \texttt{has\_socio\_demographic},
        \item \texttt{has\_cognitive\_style}, and
        \item \texttt{has\_value\_system}
    \end{enumerate}---are unchanged, giving a final set of six Boolean fields.
    \textbf{Reason}: \texttt{other\_profile\_information} was specified as free text. Free-text responses proved difficult to code consistently across reviewers and could not be aggregated for synthesis without a post hoc coding step, which would itself have been data-driven. Replacing the field with pre-specified Boolean categories places all profile-related items on a common, codable scale and allows inter-reviewer agreement to be assessed.
    \textbf{Impact}: The change narrows extraction from an open catch-all to three named categories; profile attributes falling outside the six Boolean fields are no longer captured.

    \item The searches were executed on 16 July 2026, 16 days later than the date anticipated at registration.
    \textbf{Reason}: Preparatory work took longer than anticipated.
    \textbf{Impact}: Minimal. The reported search date reflects actual execution, and the coverage window of the query (December 2022 to May 2026) was unchanged.

\end{enumerate}}

\DefineRevision{method-search-criteria}
{}
{We then applied the {\methodname} principles to studies of emergent human-like social behavior in AI systems, considering all papers published between December 2022 and May 2026.
A study was eligible if it (1) used LLMs as interacting agents in a multi-agent simulation environment; (2) examined collective or social behavior involving group dynamics, whether aggregate group-level dynamics or individual behavioral change arising from exposure to a group, rather than task-oriented benchmarks (\eg, coding, mathematics or tool use); (3) specified an explicit simulation objective or target social phenomenon; and (4) reported sufficient methodological detail, including prompts or interaction procedures, to permit assessment under our framework.
We excluded editorials, commentaries, reviews, position papers, extended abstracts, and non-English publications.

We searched Scopus, IEEE Xplore, ACM Digital Library, and arXiv, which together provide broad coverage of recent work on LLMs, multi-agent systems and AI-based social simulation across both computer science and interdisciplinary venues.
All searches were conducted on 16 July 2026 and were restricted to titles, abstracts and keywords.}

\DefineRevision{method-search-string}
{}
{The full search string was:

\texttt{("large language model*" OR LLM OR LLMs OR "generative agent*" OR "LLM-based agent*" OR "LLM agent*") AND ("social simula*" OR "large-scale simula*" OR "generative simula*" OR "large scale simula*" OR "interaction simula*" OR "interactive simula*" OR "propagation simula*" OR "social media simula*" OR "social network simula*" OR "multi-agent simula*" OR "multi agent simula*" OR "multiagent simula*" OR "social interact*" OR "multi-agent interact*" OR "multi agent interact*" OR "multiagent interact*" OR "social environment*" OR "social system*" OR "agent societ*" OR "AI societ*" OR "LLM societ*" OR ABM OR "agent based model*" OR "agent-based model*" OR "collective behavior*" OR "opinion dynamic*" OR "social dynamic*" OR "interaction dynamic*" OR "group dynamic*" OR "attitude dynamic*" OR "social influence" OR "social intelligence" OR "information propagat*" OR "collective decision*" OR homophil* OR polariz* OR polaris* OR "network formation*" OR "network grow*")}}

\DefineRevision{method-screening}
{}
{After removing duplicate records, the initial retrieval yielded 2,575 candidate records.
We then applied an AI-assisted screening procedure to exclude records falling outside the scope of the review.
Specifically, GPT-5.5~\cite{gpt55}, Claude-Sonnet-4.6~\cite{claudes46}, and Gemini-3.5-Flash~\cite{gemini35} were prompted independently to judge the eligibility of each record (see \S\ref{sec:review-prompt} for the prompt), and disagreements were resolved by majority vote across the three models.
Title-and-abstract screening retained 480 records (Fleiss' $\kappa = 0.81$; unanimous agreement, 92.0\%; mean pairwise agreement, 94.3\%).
To validate this procedure, two human reviewers independently screened a random sample of 100 records; the two reviewers agreed on 96 of 100 records (Cohen's $\kappa = 0.87$), and the remaining discrepancies were resolved by discussion.
Agreement between the model majority vote and the resolved human decision was high (Cohen's $\kappa = 0.91$; observed agreement, 97.0\%; expected agreement, 67.3\%).
Full-text screening with the same three models and prompt retained {\numpaper} papers.
Finally, a separate prompt was used to extract the individual studies reported in each paper, yielding {\numstudy} studies.}

\DefineRevision{method-data-extraction}
{}
{We operationalized the {\methodname} principles across the {\numstudy} studies using a two-stage process.
First, we extracted a set of predefined fields characterizing each simulation; all fields were either Boolean or restricted to predefined categories to ensure consistent annotation.
Second, we applied rule-based criteria to map these fields onto a binary verdict (compliance or violation) for each principle.
The extraction fields and judgment criteria for each component are detailed below.}

\DefineRevision{method-profile}
{We define \textit{Profile} heterogeneity as the presence of agent-specific attributes designed to induce statistically distinct behavioral responses to identical stimuli.
To ensure a conservative assessment in our systematic audit, we operationalized this dimension using a minimalist binary criterion rather than a measure of ``sufficiency.''
A simulation was coded as possessing the \textit{Profile} component only if agents were assigned distinct intrinsic traits—such as social status, psychometric dimensions (\eg, Big Five personality traits), or nationality.
Crucially, the mere variation of nominal identifiers (\eg, ``Agent 1'' vs. ``Agent 2'') or names was deemed insufficient to constitute a profile, as these labels lack the semantic depth required to simulate diverse human priors or social predispositions.}
{We define \textit{Profile} heterogeneity as the presence of agent-specific attributes intended to elicit statistically distinguishable behavioral responses to identical stimuli.
\rev{We coded six Boolean indicators of agent profiles, each marked true when agents were assigned: (1) socio-demographics: age, gender, occupation, education, nationality, or comparable human-background descriptors; (2) cognitive style: cognitive, perceptual, or reasoning tendencies affecting information processing, such as risk tolerance or analytical style; (3) value system: values, beliefs, ideologies, moral commitments, goals, norms, or comparable motivational orientations; (4) background story: a backstory, biography, or personal scenario; (5) personality: personality traits, character attributes, or dispositions; and (6) role: a social, occupational, or task-specific role.
A simulation was classified as compliant with the \textit{Profile} principle if at least one of the six indicators was marked true.}}

\DefineRevision{method-interaction}
{We define \textit{Interaction} as the endogenous exchange of information between autonomous agents, where an agent's output (either in its raw textual form or as a processed cognitive representation) serves as a direct stimulus for another agent's decision-making or reflection process.
To ensure methodological rigor, we strictly distinguish between genuine inter-agent communication and exogenously imposed context.
A simulation is flagged as violating the \textit{Interaction} principle if the social environment is statically constructed by the researcher rather than dynamically generated by the agent collective.
For instance, a common artifact in ``herd effect'' studies involves a prompt such as: ``All ten of your peers have chosen Option A; what is your choice?''
In this configuration, the peer behavior is an arbitrary parameter pre-determined by the experimenter, bypassing the actual social dynamics.
Under our audit, such ``pseudo-multi-agent'' designs are categorized as single-agent tasks, as they lack the reciprocal and emergent feedback loops essential for simulating collective human behavior.}
{We define \textit{Interaction} as the endogenous exchange of information between autonomous agents, whereby one agent's output---whether as raw text or as a processed environmental change---serves as a direct stimulus for another agent's decision-making or reflection.
\rev{To distinguish genuine inter-agent communication from exogenously imposed social context, we coded the interaction mechanism of each simulation into one of four categories: (1) text-based: agents directly received and processed messages or dialogue generated by other agents; (2) environment-based: agents influenced one another indirectly through changes to a shared environment or state, without direct communication; (3) mixed: both direct textual communication and environment-mediated influence were present; and (4) no interaction: one agent's response did not affect the input, state or output of any other agent.
Simulations coded as no interaction were flagged as violating the \textit{Interaction} principle.}
A common artifact in studies of herd effects, for example, is a prompt such as ``All ten of your peers have chosen Option A; what is your choice?'', in which peer behavior is a parameter fixed a priori by the experimenter and genuine social dynamics are bypassed.
Under our audit, such pseudo-multi-agent designs were classified as single-agent tasks, because they lack the reciprocal, emergent feedback loops required to simulate collective human behavior.}

\DefineRevision{method-memory}
{We operationalize \textit{Memory} as the requirement for agents to maintain temporal statefulness throughout a simulation.
A study satisfies this principle if agents possess a mechanism—whether via a dedicated memory module, an external database, or an evolving context window—to synthesize prior interactions, self-reflections, and environmental shifts into a persistent state that informs subsequent behavior.
During our audit, we distinguished between stateful and memoryless systems.
A simulation was flagged for violating the \textit{Memory} principle if agents functioned as ``pass-through'' nodes—for instance, by merely rephrasing instantaneous inputs (as observed in several ``telephone game'' studies) without evidence of longitudinal information integration or lossy compression characteristic of human cognitive processing.}
{We operationalized \textit{Memory} as the requirement that agents maintain temporal statefulness throughout a simulation.
\rev{During the audit, we first recorded whether a simulation was single-round, defined as each agent acting only once or no repeated interaction occurring across rounds.
For simulations involving repeated interaction, we further classified the memory mechanism into five categories: (1) direct storage: prior interaction content was retained verbatim in the context window; (2) retrieval-based: past information was stored externally and subsequently accessed through a retrieval or long-term memory system, which could be shared across agents; (3) state-based: memory was represented as a structured internal state rather than as raw interaction history; (4) mixed: two or more of these mechanisms were combined; and (5) no memory: agents retained neither prior interaction history nor state across rounds.
A simulation was flagged as violating the \textit{Memory} principle if it was single-round or was classified as having no memory.}}

\DefineRevision{method-minimal-control}
{Our primary assessment relied on expert manual annotation, cross-validated by an automated LLM-based audit to ensure the scalability and reproducibility of the evaluation (refer to Table~\ref{tab:review} for consolidated results).}
{To make this assessment scalable and reproducible, we annotated each simulation using six frontier LLMs (consolidated results in Table~\ref{tab:review}).
\rev{Each model received the raw prompts extracted from the simulation together with the stated simulation goal, and judged whether the prompts contained any content that could steer agents toward the behavior described in that goal.
A simulation was flagged as violating the principle when at least three of the six models identified such content.}}

\DefineRevision{method-unawareness}
{To quantify potential experimental leakage within the simulation designs, we systematically interrogated five state-of-the-art LLMs (detailed in Table~\ref{tab:unawareness-result}) by exposing them to the original experimental prompts.
We operationalized ``awareness'' as the model's capacity to both explicitly name the target social phenomenon and accurately describe its underlying mechanism.
Responses were evaluated by independent human annotators; a study was classified as violating the \textit{Unawareness} principle if a majority consensus (at least three out of five models) successfully identified the experimental intent.
This procedure assesses whether the LLM's behavior is driven by the simulated social dynamics or by demand characteristics—pre-existing knowledge of the retrieved experimental paradigm.}
{To quantify potential experimental leakage in the simulation designs, we probed six state-of-the-art LLMs (Table~\ref{tab:unawareness-result}) with the original experimental instructions, asking each model to infer the underlying simulation and its expected outcome.
We operationalized awareness as a model's capacity both to name the target social phenomenon explicitly and to describe its underlying mechanism accurately.
\rev{Three LLMs then judged whether each response correctly identified the simulation and the experimental intent.
A study was classified as violating the \textit{Unawareness} principle if at least three of the six models identified the experimental intent.
This procedure tests whether a model's behavior is driven by the simulated social dynamics or by demand characteristics, \ie, prior knowledge of the replicated experimental paradigm.}}

\DefineRevision{method-realism}
{We evaluated the \textit{Realism} of each study by examining whether its claims regarding the fidelity of AI-based simulations were anchored in empirical human data.
For a study to satisfy this criterion, any assertion that LLM-based collective behaviors replicate, provide insight into, or deviate from human society should be supported by direct comparison with real-world datasets or established empirical benchmarks.
We classified a study as failing this criterion if such claims were (1) unsubstantiated by experimental evidence, or (2) justified solely through internal comparisons with other simulated outcomes (\ie, circular validation).
For studies focused specifically on the emergence of known social phenomena (\eg, herd behavior), we required evidence that the simulation qualitatively aligns with the phenomenological characteristics of that behavior as documented in the behavioral sciences, rather than requiring new primary human data collection.}
{We assessed \textit{Realism} at the level of each study's validation strategy.
For studies concerned with the emergence of established social phenomena (\eg, herd behavior), we did not require new primary data collection; instead, we required evidence that the simulation qualitatively reproduced the phenomenological features of that behavior as documented in the behavioral sciences.
\rev{We operationalized \textit{Realism} as the type of external reference used to validate the simulation, coded into four categories: (1) human data, where simulated outcomes were compared with real-world behavioral or social data; (2) theoretical model, where simulations were evaluated against a mathematical or non-LLM agent-based model; (3) both, where empirical data and theoretical models were both used as references; and (4) none, where no external validation target was specified.
Claims that LLM-based collective behavior replicates, explains, or departs from human social behavior satisfied the \textit{Realism} principle only when supported by direct comparison with real-world datasets or established empirical benchmarks.}}

\DefineRevision{method-categorization}
{}
{To characterize variation across the reviewed literature, we classified each simulation by its primary goal and reported audit results stratified by this classification.
We first embedded each stated goal using OpenAI's \texttt{text-embedding-3-large} model and grouped semantically similar goals through hierarchical clustering.
We then manually reviewed, refined, and named the resulting clusters, and assigned each simulation to a final category.
For each category, we report the mean compliance rate for every {\methodname} component.}

\DefineRevision{method-qualitative}
{}
{We did not perform a conventional study-level risk-of-bias assessment, as the included evidence comprises heterogeneous LLM-based multi-agent simulations, to which existing tools developed for clinical, intervention, or qualitative research do not directly apply.
Instead, we appraised confidence in each major finding regarding each of the six principles using the GRADE-CERQual framework~\cite{lewin2018gradecerqual}, assessing methodological limitations, coherence, adequacy of data, and relevance (Table~\ref{tab:grade-cerqual}).}

\DefineRevision{method-review-bias}
{}
{Although we excluded studies that provided neither appendix prompts nor a link to their code, disclosure among the remaining studies was frequently still incomplete: some code URLs resolved to empty repositories, and some reported prompts were only partial.
Coding the {\numstudy} studies therefore yielded a number of \textsc{missing} fields, which we exclude from the denominator; compliance rates are computed as \textsc{true} / (\textsc{true} + \textsc{false}).
Because non-reporting plausibly signals non-compliance rather than unobserved compliance, this convention is likely to inflate the estimates, and we should therefore interpret all reported compliance rates as upper bounds.}

\DefineRevision{method-statistics-sample-size}
{}
{No statistical methods were used to pre-determined sample sizes, but our sample sizes are similar to those reported in the reproduced publications~\cite{liu2024skepticism, cisneros2024large, liu2025exploring, cho2025herd}.
The social network growth experiments~\cite{de2023emergence} are the one exception: we did not adopt the original sample size of $N = 2{,}500$.
In the original design, each arriving agent issues a single query, so cost scales as $O(N)$ and this sample size is tractable.
Enforcing \textit{Interaction} replaces that single query with conversations between the arriving agent and every existing member, raising the cost to $O(N^2)$.
We therefore used $N = 300$ with five independent runs, which preserves the qualitative dynamics of interest while remaining computationally feasible.}

\DefineRevision{method-randomization}
{}
{Because LLM outputs are non-deterministic, we repeated every experiment across independent runs.
The exception was the herd effect experiment, which already included a large number of queries, so within-run averaging provided comparable variance reduction.}

\DefineRevision{method-statistics-fake-news-propagation}
{The primary outcome measure was the population-level belief convergence rate, averaged across multiple independent runs to mitigate stochastic variance.}
{The primary outcome was belief convergence, operationalized as the final proportion of infected agents in each independent simulation run.
\rev{Each condition comprised $N_\text{run}=3$ independent runs, with the run-level proportion serving as the unit of analysis; we report means across runs.
Conditions were compared using two-sided independent-samples $t$-tests ($N = N_\text{seed} \times N_\text{run} = 18$), with mean differences reported alongside 95\% confidence intervals.}}

\DefineRevision{method-statistics-social-balance}
{}
{Each simulated relationship configuration was classified as balanced or unbalanced.
We compared the proportion of balanced configurations between the two conditions using a two-sided two-proportion $z$-test ($N = N_\text{seed} \times N_\text{run} = 640$).}

\DefineRevision{method-telephone-game-correction}
{Each simulation comprised 50 sequential turns, with each turn involving a unique agent instance tasked with relaying information to the subsequent agent.}
{\rev{Each simulation comprised $N_\text{agent}=15$ agents (therefore $N_\text{round}=15$ transmissions), with each transmission involving a unique agent instance tasked with relaying information to the subsequent agent.}}

\DefineRevision{method-statistics-telephone-game}
{}
{For statistical analysis, each chain was initialized with one of the $N_\text{seed}=20$ seed texts and passed along a chain of $N_\text{round}=15$ successive transmissions.
Every seed text was run $N_\text{run}=5$ times, giving $N = N_\text{seed} \times N_\text{run} = 100$ chains per condition ($300$ in total).
Fidelity (measured using SimCSE~\cite{gao2021simcse}) was dependent within a chain by design, as each message was generated from the preceding message in the same chain, and seed texts were reused across the three conditions.
We therefore fitted a linear mixed-effects model, fidelity $\sim$ condition $\times$ round $+\ (1 |$chain$)$.}

\DefineRevision{method-statistics-herd-effect}
{}
{The primary dependent variable was the flip rate, the proportion of trials in which the target agent abandoned its initial preference in favor of the steering target.
Trial-level flip rates were modeled using logistic regression, with the target agent's initial self-confidence, its inferred confidence in the interlocutor (our operationalization of perceived social pressure), and their interaction (flip-rate $\sim$ self-conf $+$ social-press $+$ self-conf $\times$ social-press; $N = N_\text{seed} \times N_\text{run} = 2{,}152$).}

\DefineRevision{method-statistics-social-network-growth}
{}
{We estimated the power-law exponent ($\alpha$) by maximum likelihood, choosing the lower cutoff $k_{\min}$ to minimize the Kolmogorov-Smirnov (KS) distance between the empirical and fitted cumulative distributions above the cutoff~\cite{clauset2009power}.
Goodness of fit was assessed with a parametric bootstrap KS test based on 1,000 synthetic datasets, and we report the KS statistic ($D_\text{KS}$) and the bootstrap $P$ value $P_\text{boot}$.
Following convention, we did not reject the power-law tail when $P_\text{boot} > 0.10$.}

\DefineRevision{added-floats-fig-objectives}
{}
{Four common objectives motivating AI multi-agent social simulation research. Our paper targets the two types of studies in the first row, focusing on claims about human society.}

\DefineRevision{added-floats-flow-diagram}
{}
{A flow diagram (PRISMA-compliant) describing how we conducted the systematic review.}

\DefineRevision{added-floats-table-main}
{}
{Proportion of simulations satisfying each criterion (\%). Simulations have some missing fields. We exclude them from both the numerator and the denominator, so $N$ only denotes the number of papers in a category instead of the denominator. We compute these ratios using \textsc{true} / (\textsc{true} + \textsc{false}).}

\DefineRevision{added-floats-prompt-fake-news-propagation}
{}
{The following prompt replaces [\texttt{query}] in the previous prompt for the fake news propagation simulation.}

\DefineRevision{added-floats-prompt-social-balance}
{}
{The following prompt replaces [\texttt{query}] in the previous prompt for the social balance simulation.}

\DefineRevision{added-floats-prompt-telephone-game}
{}
{The following prompt replaces [\texttt{query}] in the previous prompt for the telephone game simulation.}

\DefineRevision{added-floats-prompt-herd-effect}
{}
{The following prompt replaces [\texttt{query}] in the previous prompt for the herd effect simulation.}

\DefineRevision{added-floats-prompt-social-network-growth}
{}
{The following prompt replaces [\texttt{query}] in the previous prompt for the social network growth simulation.}

\DefineRevision{added-floats-taxonomy-1}
{}
{The taxonomy of simulation goals (category 1). $N$ is the number of simulations assigned to each category.}

\DefineRevision{added-floats-taxonomy-2}
{}
{The taxonomy of simulation goals (category 2). $N$ is the number of simulations assigned to each category.}

\DefineRevision{added-floats-taxonomy-3}
{}
{The taxonomy of simulation goals (category 3). $N$ is the number of simulations assigned to each category.}

\DefineRevision{added-floats-taxonomy-4}
{}
{The taxonomy of simulation goals (category 4). $N$ is the number of simulations assigned to each category.}

\DefineRevision{added-floats-taxonomy-5}
{}
{The taxonomy of simulation goals (category 5). $N$ is the number of simulations assigned to each category.}

\DefineRevision{added-floats-claims}
{}
{Claims of emergent or human-consistent collective behavior in the five reproduced studies. Page numbers refer to the publicly available versions of the papers.}

\DefineRevision{added-floats-grade-cerqual}
{}
{GRADE-CERQual evidence profile for the review findings. Each finding is rated for methodological limitations, coherence, adequacy, and relevance, which are combined into an overall assessment of confidence in the evidence.}

%% file: Sections/0_Abstract.tex
Large language models (LLMs) are increasingly used to simulate human collective behavior, yet claims that such simulations are human-like remain largely untested.
We conducted a systematic audit (pre-registered on OSF) of LLM-based social simulations across four databases (Scopus, IEEE Xplore, ACM Digital Library, and arXiv).
Across {\numstudy} studies reported in {\numpaper} recent papers, we applied six methodological evaluations: agent Profile, Interaction, Memory, Minimal-Control, Unawareness, and Realism ({\methodname}).
Coding every study against pre-specified rules, we revealed that PIM were met more often than MUR.
Frontier LLMs correctly identified the underlying social experiment in {\avgU} of cases, and {\avgM} of prompts imposed constraints that pre-determined the outcome.
These compliance rates are upper bounds, because incomplete methodological reporting (for example, unreleased prompts) limits the available evidence.
Reproducing five representative experiments (\eg, opinion dynamics), we found that reported collective phenomena often vanish or reverse once {\methodname} principles are enforced, indicating that many ``emergent'' behaviors are methodological artifacts rather than genuine social dynamics.
Current LLM simulations may therefore capture model-specific biases rather than universal features of human social behavior, raising concerns about their use as scientific proxies for human society.

%% file: Sections/1_Introduction.tex
\section{Introduction}

\begin{figure}[t]
    \centering
    \includegraphics[width=1.0\linewidth]{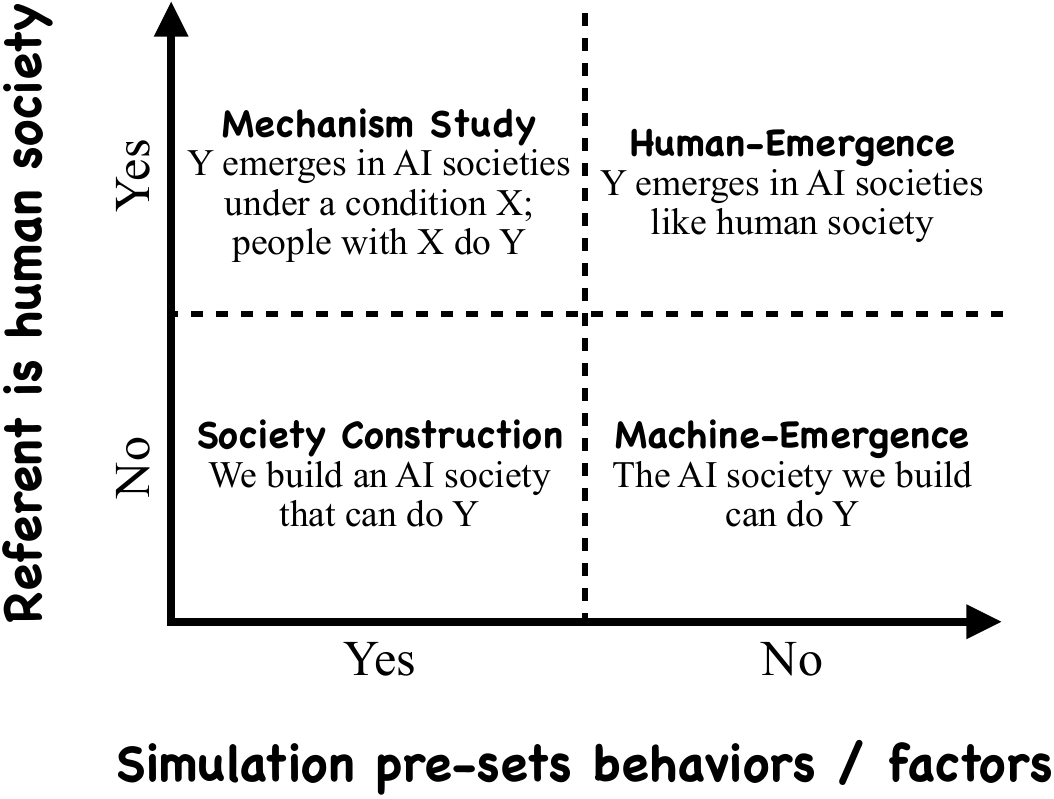}
    \caption{\InsertRevision{added-floats-fig-objectives}}
    \label{fig:scope}
\end{figure}

The rapid evolution of large language models (LLMs) in reasoning, communication, and coordination~\cite{huang2025competing, tran2025multi, agashe2025llm} has positioned them as potential ``digital twins'' for human behavior.
This has birthed a burgeoning field of LLM-based multi-agent social simulation (MASS), where researchers deploy ``AI societies'' to simulate complex collective phenomena, such as cooperation, polarization, information diffusion, or network formation~\cite{mou2024unveiling, zhou2025sotopia, zhang2024exploring, liu2025exploring}.
\InsertRevision{intro-define-scope}

However, the scientific utility of these simulations hinges on a fundamental assumption: that the observed collective patterns are truly emergent---arising from the intrinsic interactions of autonomous agents---rather than being encoded by the experimenter.
In human behavioral research, it is well established that experimental validity is often compromised by ``experimenter visibility effects.''
These include \textbf{demand characteristics}, where participants subconsciously alter their behavior to fit the perceived experimental goal~\cite{orne2017social}; \textbf{the Hawthorne effect}, where the mere awareness of observation induces behavioral change~\cite{adair1984hawthorne, mayo2004human, roethlisberger2003management}; and \textbf{social desirability bias}, the tendency to project normative rather than authentic responses, such as overstating prosocial behavior~\cite{grimm2010social} or downplaying controversial opinions~\cite{nederhof1985methods}.
As recent evidence suggests that LLMs are not immune to these psychological artifacts~\cite{shapira2024clever, shao2025spurious, cheng2025social}, the validity of current MASS studies remains deeply uncertain~\cite{anthis2025llm}.
\InsertRevision{intro-current-drawbacks}

\InsertRevision{intro-inductive-deductive}
\begin{enumerate}[noitemsep]
    \item \textit{\textbf{P}rofile} (Homogeneity): A lack of agent diversity, where identical model weights lead to artificial behavioral convergence.
    \item \textit{\textbf{I}nteraction} (Pseudo-Agency): Simulations that reduce multi-agent dynamics to sequential single-agent reasoning, stripping agents of true social agency.
    \item \textit{\textbf{M}emory} (Statelessness): The absence of persistent, ``digested'' internal states, preventing the formation of authentic social identities or belief evolution.
    \item \textit{\textbf{M}inimal-Control} (Goal-Injection): The use of ``lead-the-witness'' prompting that explicitly instructs agents to exhibit specific biases (\eg, confirmation bias), pre-determining the outcome.
    \item \textit{\textbf{U}nawareness} (Visibility Leakage): Experimental designs that are so transparent that the LLM can identify the underlying social theory, triggering social desirability or ``Clever Hans'' effects.\footnote{The term originates from a German horse in the early 1900s that seemed to solve math problems but was actually responding to subtle, unconscious cues from its trainer.}
    \item \textit{\textbf{R}ealism} (Circular Validation): A reliance on idealized mathematical models as ground truth rather than empirical human data, creating a ``model-to-model'' circularity that lacks ecological validity.
\end{enumerate}

\InsertRevision{intro-systematic-review}
Our audit revealed that {\numfail} of existing studies violate at least one {\methodname} principle, with a majority failing to safeguard against experimenter visibility.
Finally, through the reproduction of five experiments, we demonstrated that many claimed ``emergent'' behaviors---such as the ``telephone game'' message decay or social network balance---either vanish or fundamentally reverse when {\methodname} principles are enforced.
The flawed experimental designs invalidate the target social experiments by ignoring key human-like features, thus producing conclusions that are often unreliable.
These findings suggest that much of what is currently interpreted as ``emergent AI social behavior'' is, in fact, a methodological artifact.
We argue that for MASS to serve as a legitimate prediction of human behavior, the field must transition from ``prompt engineering'' to a more rigorous ``behavioral engineering'' paradigm.

%% file: Sections/2_Results.tex
\section{Results}

\subsection{Taxonomy of Methodological Flaws}
\label{sec:principle}

\InsertRevision{results-inductive-deductive}

\begin{AIbox}{230pt}{Definition: \textbf{Profile}}
{
Agents should possess diverse backgrounds, such as socio-demographics, cognitive styles, or value systems, to ensure systemic heterogeneity, avoiding the artifacts of a monolithic model distribution.
}
\end{AIbox}
Homogeneity remains a critical threat to the validity of AI societies.
When agents are merely copies of a single LLM backbone without explicit diversification, the system risks ``collapsing'' into a consensus that reflects model bias rather than social dynamics~\cite{reid2025risk, xiao2026chameleon, pitre2025consensagent, huang2026knowing, hegazy2024diversity}.
In contrast, personality is an inherent human attribute rather than an optional modeling choice.
Recent studies have demonstrated that persona-prompting can steer LLM behaviors~\cite{chen2024persona, wang2024incharacter, huang2024humanity, wang2025coser, huang2024apathetic}.
To achieve \textbf{ecological validity}, simulations must move beyond default settings like ``a helpful assistant'' and incorporate diverse model architectures or deeply grounded personas to capture how individual variances shape collective outcomes.

\begin{AIbox}{230pt}{Definition: \textbf{Interaction}}
{
Agents should exert agency via direct or indirect communication, reacting to the specific actions of others or the environment shaped by others rather than researcher-provided statistical aggregates.
}
\end{AIbox}
Many ``multi-agent'' claims are undermined by \textbf{pseudo-interaction} designs, relying on single-agent formulations.
For instance, studies on herd effects often reduce social complexity to a single-agent decision task by embedding aggregated peer data (\eg, ``10 others chose A'') directly into the prompt~\cite{cho2025herd, liu2025exploring}.
Similarly, in social network growth~\cite{de2023emergence}, a single agent is provided only with others' friend counts to make decisions.
While such simplifications make simulation more tractable, their conclusions are difficult to generalize to real-world scenarios, where dynamic interactions among agents or changes to the environment crucially shape individual behaviors~\cite{ng2026social, ran2025bookworld, huang2025resilience, huang2025competing}.
Authentic emergence requires that behaviors arise from the ``ground up'' through decentralized interactions, where agents perceive and modify a shared environment.

\begin{AIbox}{230pt}{Definition: \textbf{Memory}}
{
Agents should maintain and update persistent internal states across time, allowing information to be internalized, retained, and re-expressed rather than rephrased statelessly.
}
\end{AIbox}
Social phenomena such as the ``telephone game'' or rumor propagation rely on the \textbf{lossy compression} and subjective interpretation of information~\cite{hirst2012remembering}.
However, current simulations often treat memory as a static buffer, instructing agents to merely paraphrase messages~\cite{liu2025exploring, perez2025llms}.
This neglects the crucial role of information internalization.
Without a persistent memory architecture that allows for the integration of new data with existing beliefs, AI agents fail to replicate the cognitive realism necessary for long-term social evolution~\cite{wang2025limits}.

\begin{AIbox}{230pt}{Definition: \textbf{Minimal-Control}}
{
Agents should be provided only the essential environmental context and action space, minimizing demand characteristics. Observed collective behaviors must emerge from agents themselves rather than from researcher-imposed behavioral cues.
}
\end{AIbox}
Prompt engineering has been shown to be effective for a wide range of LLM-based downstream applications~\cite{sahoo2024systematic, schulhoff2024prompt}.
However, this approach often introduces researcher bias into social simulations.
For example, to examine belief convergence in fake news propagation, \citet{liu2024skepticism} explicitly instruct agents to ``demonstrate confirmation bias.''
Researchers risk measuring the model's instruction-following abilities rather than its intrinsic social tendencies.
To ensure the \textbf{construct validity} of emergent phenomena, prompts must remain neutral.
If the simulation's outcome is pre-determined by the system prompt, the ``social behavior'' observed is a methodological artifact rather than a scientific discovery.
\InsertRevision{results-profile-mincontrol}

\begin{figure*}[t]
    \centering
    \includegraphics[width=0.8\linewidth]{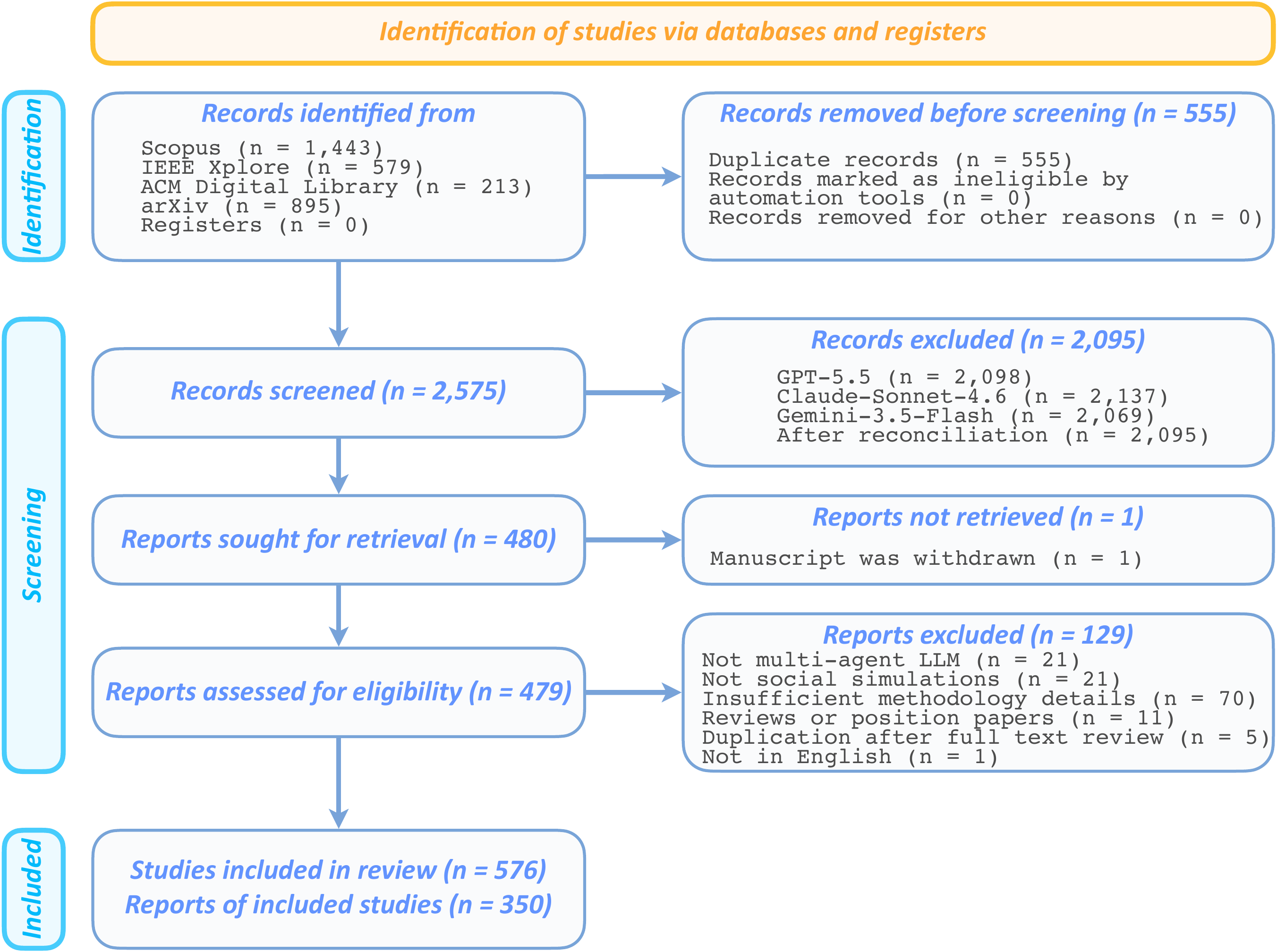}
    \caption{\InsertRevision{added-floats-flow-diagram}}
    \label{fig:flow-diagram}
\end{figure*}

\begin{table*}[t]
\centering
\caption{\InsertRevision{added-floats-table-main}}
\label{tab:review}
\resizebox{1.0\linewidth}{!}{
\rowcolors{2}{blue!10}{white}
\begin{tabular}{lccccccc}
\toprule
Category & $N$ & \bf Profile & \bf Interaction & \bf Memory & \bf Min.-Control & \bf Unawareness & \bf Realism \\
\midrule
\textbf{1. Strategic Interaction and Social Dilemmas} & 104 & 80.2 & 92.3 & 86.8 & 45.7 & 17.3 & 32.3 \\
\quad 1.1 Cooperation, reciprocity, and trust & 29 & 75.9 & 96.6 & 92.3 & 54.2 & 29.2 & 36.0 \\
\quad 1.2 Public goods and common-pool resources & 28 & 80.0 & 96.4 & 95.5 & 57.9 & 10.5 & 29.6 \\
\quad 1.3 Bargaining over division; ultimatum / dictator games & 16 & 81.2 & 68.8 & 53.3 & 57.1 & 0.0 & 56.2 \\
\quad 1.4 Coordination games & 8 & 50.0 & 87.5 & 75.0 & 60.0 & 40.0 & 25.0 \\
\quad 1.5 Social deduction and hidden-role games & 18 & 100.0 & 100.0 & 100.0 & 13.3 & 6.7 & 22.2 \\
\quad 1.6 Zero-sum and pure-competition games & 5 & 80.0 & 100.0 & 100.0 & 0.0 & 50.0 & 0.0 \\
\midrule
\textbf{2. Opinion and Information Dynamics} & 196 & 81.6 & 99.5 & 78.5 & 47.4 & 30.4 & 42.2 \\
\quad 2.1 Opinion dynamics and attitude change & 57 & 75.0 & 100.0 & 90.0 & 53.1 & 30.6 & 38.2 \\
\quad 2.2 Conformity and peer influence & 34 & 76.5 & 97.1 & 58.1 & 59.4 & 15.6 & 17.6 \\
\quad 2.3 Polarization, echo chambers, and information cocoons & 35 & 84.8 & 100.0 & 95.2 & 46.4 & 46.4 & 58.1 \\
\quad 2.4 Persuasion and debate & 18 & 94.4 & 100.0 & 87.5 & 5.9 & 23.5 & 38.9 \\
\quad 2.5 Information diffusion and misinformation & 31 & 89.3 & 100.0 & 57.1 & 34.6 & 30.8 & 69.0 \\
\quad 2.6 Network formation and influence structure & 21 & 81.0 & 100.0 & 78.9 & 68.4 & 36.8 & 35.0 \\
\midrule
\textbf{3. Interpersonal and Group Processes} & 131 & 95.3 & 93.0 & 90.3 & 43.9 & 50.5 & 31.5 \\
\quad 3.1 Negotiation & 26 & 96.2 & 100.0 & 96.0 & 52.2 & 52.2 & 15.4 \\
\quad 3.2 Collaboration and division of labor & 22 & 90.9 & 95.5 & 94.4 & 27.8 & 61.1 & 14.3 \\
\quad 3.3 Group deliberation and collective choice & 22 & 90.9 & 81.0 & 80.0 & 52.9 & 70.6 & 47.6 \\
\quad 3.4 Social norms, compliance, and sanctioning & 13 & 90.9 & 75.0 & 55.6 & 50.0 & 50.0 & 8.3 \\
\quad 3.5 Deception, manipulation, and trust violation & 19 & 100.0 & 100.0 & 100.0 & 27.8 & 27.8 & 84.2 \\
\quad 3.6 Emotion, relationships, and social support & 8 & 100.0 & 100.0 & 100.0 & 66.7 & 33.3 & 57.1 \\
\quad 3.7 Self-disclosure, identity, and status & 8 & 100.0 & 100.0 & 100.0 & 66.7 & 66.7 & 0.0 \\
\quad 3.8 General goal-directed social interaction & 13 & 100.0 & 92.3 & 90.9 & 41.7 & 41.7 & 15.4 \\
\midrule
\textbf{4. Collective Behavior and Macro-level Emergence} & 96 & 91.7 & 94.7 & 90.8 & 47.9 & 45.2 & 37.6 \\
\quad 4.1 Market and price emergence & 33 & 96.3 & 100.0 & 95.7 & 36.7 & 36.7 & 39.4 \\
\quad 4.2 Crowd movement and spatial distribution & 10 & 100.0 & 90.0 & 100.0 & 50.0 & 100.0 & 60.0 \\
\quad 4.3 Evacuation and collective behavior under crisis & 16 & 100.0 & 93.3 & 88.9 & 63.6 & 27.3 & 62.5 \\
\quad 4.4 Emergence of norms, institutions, and culture & 27 & 76.0 & 88.9 & 81.0 & 60.0 & 40.0 & 24.0 \\
\quad 4.5 Open-ended agent societies & 10 & 100.0 & 100.0 & 100.0 & 37.5 & 87.5 & 0.0 \\
\midrule
\textbf{5. Social Cognition} & 49 & 73.3 & 60.4 & 60.6 & 51.2 & 60.5 & 29.2 \\
\quad 5.1 Cognitive bias in social contexts & 18 & 62.5 & 77.8 & 50.0 & 50.0 & 64.3 & 38.9 \\
\quad 5.2 Stereotypes and social bias & 6 & 83.3 & 100.0 & 100.0 & 33.3 & 50.0 & 0.0 \\
\quad 5.3 Individual-level psychology and behavioral realism & 25 & 78.3 & 37.5 & 52.6 & 56.5 & 60.9 & 29.2 \\
\midrule
\textbf{All} & 576 & 85.4 & 92.6 & 83.5 & 46.7 & 37.7 & 36.1 \\
\bottomrule
\end{tabular}
}
\end{table*}

\begin{AIbox}{230pt}{Definition: \textbf{Unawareness}}
{
Agents should remain unaware of the experimental hypothesis, design, and evaluation criteria. This reduces experimental biases, where models adjust their behavior to align with perceived social expectations or experimental goals.
}
\end{AIbox}
LLMs are trained on vast corpora containing classic social science experiments (\eg, Sugarscape, Prisoner's Dilemma).
This creates a risk of \textbf{epistemic leakage}: if a model recognizes the experimental setup, its response may be biased by ``social desirability'' or ``the Hawthorne effect''---the tendency to perform according to a recognized theory.
In contrast to \textit{Minimal-Control}, which primarily concerns the instruction design, \textit{Unawareness} pertains to the models themselves.
Even a carefully designed minimal-control instruction can still reflect a classical experimental setup so famous that models can easily infer it, such as \citet{masumori2025large}'s Sugarscape experiment and \citet{cau2025selective}'s studies on opinion dynamics.
We argue that frontier LLMs can be ``too intelligent'' for certain social experiments.
Researchers must audit whether models can infer the study's objective, as meta-awareness systematically invalidates behavioral authenticity.

\begin{AIbox}{230pt}{Definition: \textbf{Realism}}
{
Simulations should use empirical data from real-world human societies as references rather than simplified theoretical models when making claims about the similarity of AI's emergent behaviors and real-world human dynamics.
}
\end{AIbox}
Social science has long utilized simplified mathematical models, such as Heider's balance theory~\cite{heider1946attitudes} and the Sugarscape experiment~\cite{epstein1996growing}, to explain complex human societies.
A significant portion of current MASS research evaluates LLMs solely on their ability to replicate these idealized mathematical outcomes~\cite{de2023emergence, cisneros2024large, borah2025mind}.
While theoretical models provide valuable mechanistic insights, they should not be treated as the sole ground truth for validating AI social intelligence~\cite{larooij2025large}.
We contend that while theoretical alignment is a necessary first step, it is insufficient for claiming that LLMs simulate human behaviors.
To bridge the gap between theory and reality, simulations must be grounded in and validated against high-fidelity empirical traces (\eg, social media logs, historical data), ensuring that AI societies reflect the nuances of actual human collectives.

\begin{table*}[t]
\centering
\small
\caption{Proportion of simulations judged to satisfy \textbf{Unawareness} by each model (\%). The model versions are: GPT-5.6-Terra, Claude-Sonnet-5, Gemini-3.7-Flash, DeepSeek-V4-Pro, Qwen-3.8-2.4T-A95B, and Kimi-K3.}
\label{tab:unawareness-result}
\resizebox{1.0\linewidth}{!}{
\rowcolors{2}{blue!10}{white}
\begin{tabular}{lcccccccc}
\toprule
Category & $N$ & \bf GPT-5.6 & \bf Claude-5 & \bf Gemini-3.7 & \bf DeepSeek-V4 & \bf Qwen3.8 & \bf Kimi-K3 & \bf All \\
\midrule
\textbf{1. Strategic Interaction and Social Dilemmas} & 104 & 16.0 & 19.8 & 12.3 & 12.3 & 21.0 & 13.6 & 15.8 \\
\quad 1.1 Cooperation, reciprocity, and trust & 29 & 20.8 & 29.2 & 16.7 & 12.5 & 37.5 & 20.8 & 22.9 \\
\quad 1.2 Public goods and common-pool resources & 28 & 10.5 & 10.5 & 5.3 & 5.3 & 10.5 & 10.5 & 8.8 \\
\quad 1.3 Bargaining over division; ultimatum / dictator games & 16 & 7.1 & 7.1 & 7.1 & 0.0 & 7.1 & 0.0 & 4.8 \\
\quad 1.4 Coordination games & 8 & 40.0 & 40.0 & 40.0 & 40.0 & 20.0 & 40.0 & 36.7 \\
\quad 1.5 Social deduction and hidden-role games & 18 & 6.7 & 13.3 & 0.0 & 13.3 & 20.0 & 0.0 & 8.9 \\
\quad 1.6 Zero-sum and pure-competition games & 5 & 50.0 & 50.0 & 50.0 & 50.0 & 25.0 & 50.0 & 45.8 \\
\midrule
\textbf{2. Opinion and Information Dynamics} & 196 & 27.5 & 36.3 & 23.4 & 40.4 & 28.7 & 25.7 & 30.3 \\
\quad 2.1 Opinion dynamics and attitude change & 57 & 30.6 & 40.8 & 30.6 & 49.0 & 30.6 & 24.5 & 34.4 \\
\quad 2.2 Conformity and peer influence & 34 & 9.4 & 21.9 & 15.6 & 18.8 & 18.8 & 12.5 & 16.1 \\
\quad 2.3 Polarization, echo chambers, and information cocoons & 35 & 35.7 & 39.3 & 25.0 & 53.6 & 32.1 & 35.7 & 36.9 \\
\quad 2.4 Persuasion and debate & 18 & 23.5 & 47.1 & 17.6 & 35.3 & 29.4 & 23.5 & 29.4 \\
\quad 2.5 Information diffusion and misinformation & 31 & 23.1 & 34.6 & 11.5 & 42.3 & 26.9 & 23.1 & 26.9 \\
\quad 2.6 Network formation and influence structure & 21 & 47.4 & 36.8 & 36.8 & 36.8 & 36.8 & 42.1 & 39.5 \\
\midrule
\textbf{3. Interpersonal and Group Processes} & 131 & 45.8 & 45.8 & 33.6 & 43.9 & 52.3 & 37.4 & 43.1 \\
\quad 3.1 Negotiation & 26 & 39.1 & 52.2 & 34.8 & 52.2 & 52.2 & 43.5 & 45.7 \\
\quad 3.2 Collaboration and division of labor & 22 & 61.1 & 50.0 & 38.9 & 44.4 & 72.2 & 50.0 & 52.8 \\
\quad 3.3 Group deliberation and collective choice & 22 & 58.8 & 52.9 & 41.2 & 52.9 & 76.5 & 52.9 & 55.9 \\
\quad 3.4 Social norms, compliance, and sanctioning & 13 & 30.0 & 50.0 & 40.0 & 40.0 & 50.0 & 30.0 & 40.0 \\
\quad 3.5 Deception, manipulation, and trust violation & 19 & 38.9 & 33.3 & 16.7 & 27.8 & 22.2 & 16.7 & 25.9 \\
\quad 3.6 Emotion, relationships, and social support & 8 & 33.3 & 33.3 & 16.7 & 33.3 & 33.3 & 16.7 & 27.8 \\
\quad 3.7 Self-disclosure, identity, and status & 8 & 66.7 & 66.7 & 66.7 & 66.7 & 66.7 & 66.7 & 66.7 \\
\quad 3.8 General goal-directed social interaction & 13 & 41.7 & 33.3 & 33.3 & 41.7 & 41.7 & 25.0 & 36.1 \\
\midrule
\textbf{4. Collective Behavior and Macro-level Emergence} & 96 & 46.6 & 50.7 & 37.0 & 46.6 & 46.6 & 37.0 & 44.1 \\
\quad 4.1 Market and price emergence & 33 & 36.7 & 36.7 & 33.3 & 43.3 & 46.7 & 30.0 & 37.8 \\
\quad 4.2 Crowd movement and spatial distribution & 10 & 100.0 & 100.0 & 75.0 & 100.0 & 100.0 & 75.0 & 91.7 \\
\quad 4.3 Evacuation and collective behavior under crisis & 16 & 36.4 & 36.4 & 36.4 & 27.3 & 27.3 & 36.4 & 33.3 \\
\quad 4.4 Emergence of norms, institutions, and culture & 27 & 40.0 & 60.0 & 30.0 & 35.0 & 35.0 & 25.0 & 37.5 \\
\quad 4.5 Open-ended agent societies & 10 & 87.5 & 75.0 & 50.0 & 87.5 & 75.0 & 75.0 & 75.0 \\
\midrule
\textbf{5. Social Cognition} & 49 & 62.8 & 55.8 & 46.5 & 62.8 & 48.8 & 37.2 & 52.3 \\
\quad 5.1 Cognitive bias in social contexts & 18 & 64.3 & 57.1 & 57.1 & 64.3 & 50.0 & 50.0 & 57.1 \\
\quad 5.2 Stereotypes and social bias & 6 & 66.7 & 50.0 & 33.3 & 50.0 & 50.0 & 33.3 & 47.2 \\
\quad 5.3 Individual-level psychology and behavioral realism & 25 & 60.9 & 56.5 & 43.5 & 65.2 & 47.8 & 30.4 & 50.7 \\
\midrule
\textbf{All} & 576 & 35.8 & 39.6 & 28.0 & 39.4 & 37.3 & 29.1 & 34.8 \\
\bottomrule
\end{tabular}
}
\end{table*}

\begin{table*}[t]
\centering
\caption{Proportion of simulations judged to satisfy \textbf{Minimal-Control} by each model (\%). The model versions are: GPT-5.6-Terra, Claude-Sonnet-5, Gemini-3.7-Flash, DeepSeek-V4-Pro, Qwen-3.8-2.4T-A95B, and Kimi-K3.}
\label{tab:minimal-control-result}
\resizebox{1.0\linewidth}{!}{
\rowcolors{2}{blue!10}{white}
\begin{tabular}{lcccccccc}
\toprule
Category & $N$ & \bf GPT-5.6 & \bf Claude-5 & \bf Gemini-3.7 & \bf DeepSeek-V4 & \bf Qwen3.8 & \bf Kimi-K3 & \bf All \\
\midrule
\textbf{1. Strategic Interaction and Social Dilemmas} & 104 & 32.1 & 42.0 & 63.0 & 50.6 & 49.4 & 51.9 & 48.1 \\
\quad 1.1 Cooperation, reciprocity, and trust & 29 & 37.5 & 54.2 & 75.0 & 66.7 & 58.3 & 62.5 & 59.0 \\
\quad 1.2 Public goods and common-pool resources & 28 & 52.6 & 36.8 & 73.7 & 57.9 & 63.2 & 57.9 & 57.0 \\
\quad 1.3 Bargaining over division; ultimatum / dictator games & 16 & 42.9 & 50.0 & 57.1 & 64.3 & 57.1 & 57.1 & 54.8 \\
\quad 1.4 Coordination games & 8 & 20.0 & 60.0 & 80.0 & 80.0 & 60.0 & 80.0 & 63.3 \\
\quad 1.5 Social deduction and hidden-role games & 18 & 0.0 & 26.7 & 46.7 & 6.7 & 20.0 & 26.7 & 21.1 \\
\quad 1.6 Zero-sum and pure-competition games & 5 & 0.0 & 0.0 & 0.0 & 0.0 & 0.0 & 0.0 & 0.0 \\
\midrule
\textbf{2. Opinion and Information Dynamics} & 196 & 28.7 & 40.4 & 75.4 & 50.3 & 51.5 & 58.5 & 50.8 \\
\quad 2.1 Opinion dynamics and attitude change & 57 & 26.5 & 46.9 & 73.5 & 57.1 & 55.1 & 69.4 & 54.8 \\
\quad 2.2 Conformity and peer influence & 34 & 50.0 & 37.5 & 84.4 & 62.5 & 62.5 & 65.6 & 60.4 \\
\quad 2.3 Polarization, echo chambers, and information cocoons & 35 & 17.9 & 39.3 & 71.4 & 46.4 & 53.6 & 53.6 & 47.0 \\
\quad 2.4 Persuasion and debate & 18 & 5.9 & 23.5 & 82.4 & 11.8 & 11.8 & 17.6 & 25.5 \\
\quad 2.5 Information diffusion and misinformation & 31 & 19.2 & 38.5 & 61.5 & 38.5 & 38.5 & 46.2 & 40.4 \\
\quad 2.6 Network formation and influence structure & 21 & 47.4 & 47.4 & 84.2 & 68.4 & 73.7 & 78.9 & 66.7 \\
\midrule
\textbf{3. Interpersonal and Group Processes} & 131 & 26.2 & 42.1 & 74.8 & 47.7 & 43.0 & 51.4 & 47.5 \\
\quad 3.1 Negotiation & 26 & 26.1 & 47.8 & 78.3 & 65.2 & 43.5 & 65.2 & 54.3 \\
\quad 3.2 Collaboration and division of labor & 22 & 5.6 & 33.3 & 50.0 & 38.9 & 33.3 & 27.8 & 31.5 \\
\quad 3.3 Group deliberation and collective choice & 22 & 41.2 & 41.2 & 82.4 & 58.8 & 52.9 & 64.7 & 56.9 \\
\quad 3.4 Social norms, compliance, and sanctioning & 13 & 40.0 & 50.0 & 60.0 & 40.0 & 40.0 & 50.0 & 46.7 \\
\quad 3.5 Deception, manipulation, and trust violation & 19 & 16.7 & 27.8 & 83.3 & 27.8 & 27.8 & 27.8 & 35.2 \\
\quad 3.6 Emotion, relationships, and social support & 8 & 66.7 & 66.7 & 100.0 & 66.7 & 66.7 & 66.7 & 72.2 \\
\quad 3.7 Self-disclosure, identity, and status & 8 & 33.3 & 66.7 & 66.7 & 66.7 & 66.7 & 66.7 & 61.1 \\
\quad 3.8 General goal-directed social interaction & 13 & 16.7 & 41.7 & 83.3 & 33.3 & 50.0 & 66.7 & 48.6 \\
\midrule
\textbf{4. Collective Behavior and Macro-level Emergence} & 96 & 20.5 & 45.2 & 74.0 & 56.2 & 46.6 & 52.1 & 49.1 \\
\quad 4.1 Market and price emergence & 33 & 13.3 & 40.0 & 70.0 & 50.0 & 36.7 & 43.3 & 42.2 \\
\quad 4.2 Crowd movement and spatial distribution & 10 & 50.0 & 75.0 & 100.0 & 75.0 & 50.0 & 75.0 & 70.8 \\
\quad 4.3 Evacuation and collective behavior under crisis & 16 & 36.4 & 54.5 & 90.9 & 54.5 & 63.6 & 63.6 & 60.6 \\
\quad 4.4 Emergence of norms, institutions, and culture & 27 & 15.0 & 40.0 & 70.0 & 65.0 & 55.0 & 60.0 & 50.8 \\
\quad 4.5 Open-ended agent societies & 10 & 25.0 & 50.0 & 62.5 & 50.0 & 37.5 & 37.5 & 43.8 \\
\midrule
\textbf{5. Social Cognition} & 49 & 34.9 & 37.2 & 76.7 & 53.5 & 46.5 & 60.5 & 51.6 \\
\quad 5.1 Cognitive bias in social contexts & 18 & 21.4 & 14.3 & 85.7 & 50.0 & 35.7 & 57.1 & 44.0 \\
\quad 5.2 Stereotypes and social bias & 6 & 33.3 & 16.7 & 66.7 & 33.3 & 33.3 & 33.3 & 36.1 \\
\quad 5.3 Individual-level psychology and behavioral realism & 25 & 43.5 & 56.5 & 73.9 & 60.9 & 56.5 & 69.6 & 60.1 \\
\midrule
\textbf{All} & 576 & 28.0 & 41.5 & 73.1 & 50.9 & 48.0 & 54.9 & 49.4 \\
\bottomrule
\end{tabular}
}
\end{table*}

\subsection{Systematic Audit of LLM Social Simulations}
\label{sec:review}

\paragraph{Prevalence of Methodological Vulnerabilities.}
\InsertRevision{results-systematic-audit}

\paragraph{Quantifying Experimental Leakage: The Unawareness Test.}
The \textit{Unawareness} principle is critical to mitigating experimental biases.
By definition, it assesses whether LLMs possess knowledge of the overall experimental context; the evaluation is therefore to query the models explicitly instead of relying on human annotations.
We devised a special prompt (\S\ref{sec:mc-u-prompt}) that requires models to infer the simulation goal solely from the given instructions from all {\numstudy} studies.
\InsertRevision{results-unawareness}
\InsertRevision{results-exp-awareness-citation}

\begin{figure*}[t]
  \subfloat[Num. of \textit{Skeptical} agents.]{
    \includegraphics[width=0.32\linewidth]{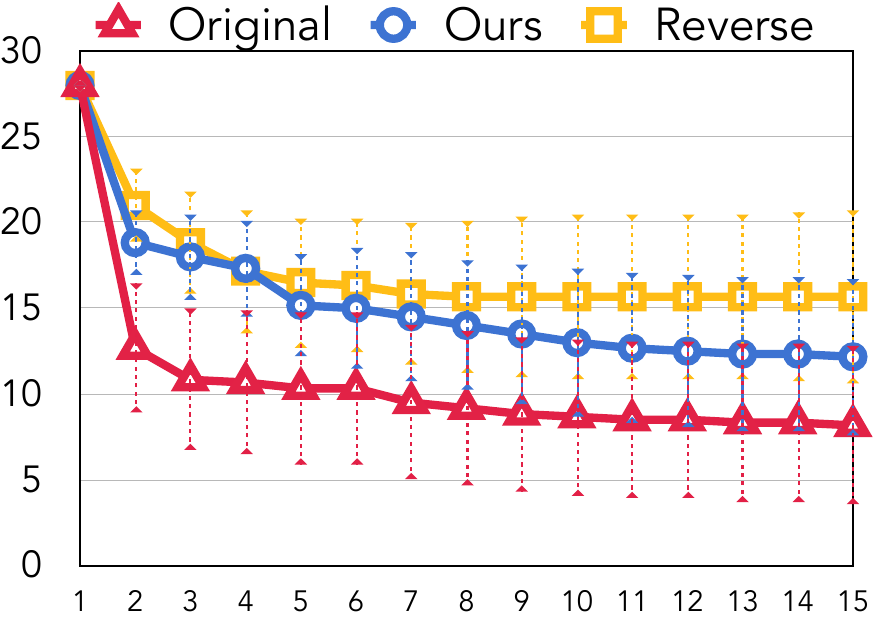}
    \label{fig:fake-news-skeptical}
  }
  \subfloat[Num. of \textit{Infected} agents.]{
    \includegraphics[width=0.32\linewidth]{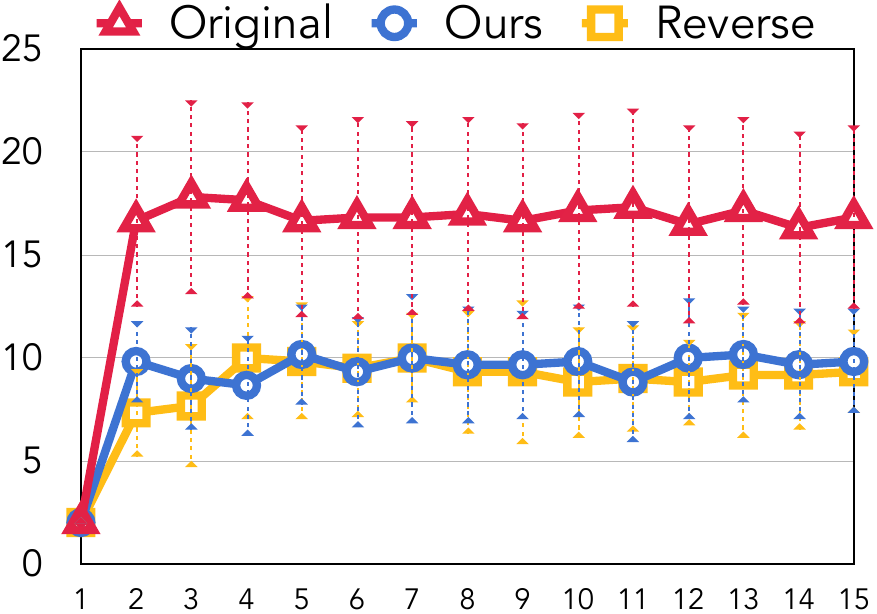}
    \label{fig:fake-news-infected}
  }
  \subfloat[Num. of \textit{Recovered} agents.]{
    \includegraphics[width=0.32\linewidth]{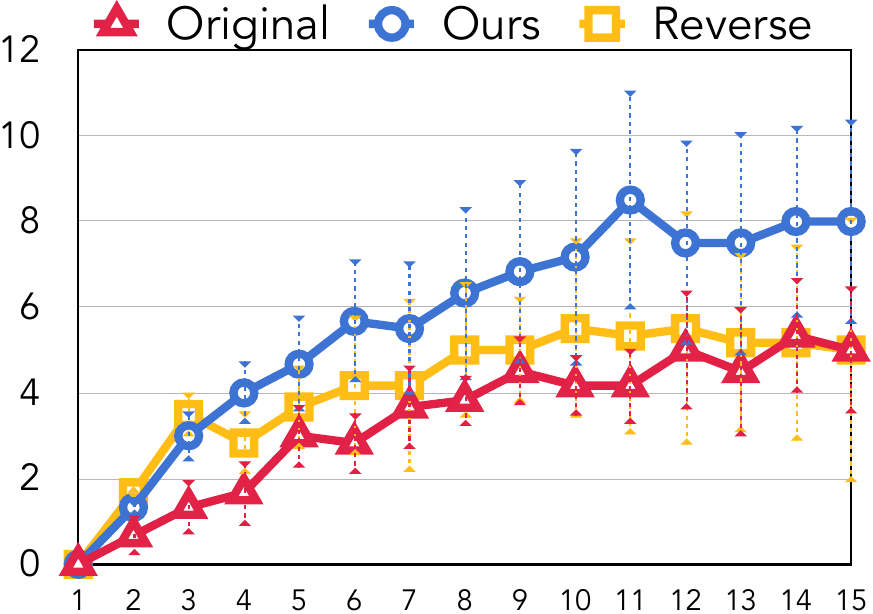}
    \label{fig:fake-news-recovered}
  }
  \\
  \subfloat[Distribution of \textit{All} relationships.]{
    \includegraphics[width=0.32\linewidth]{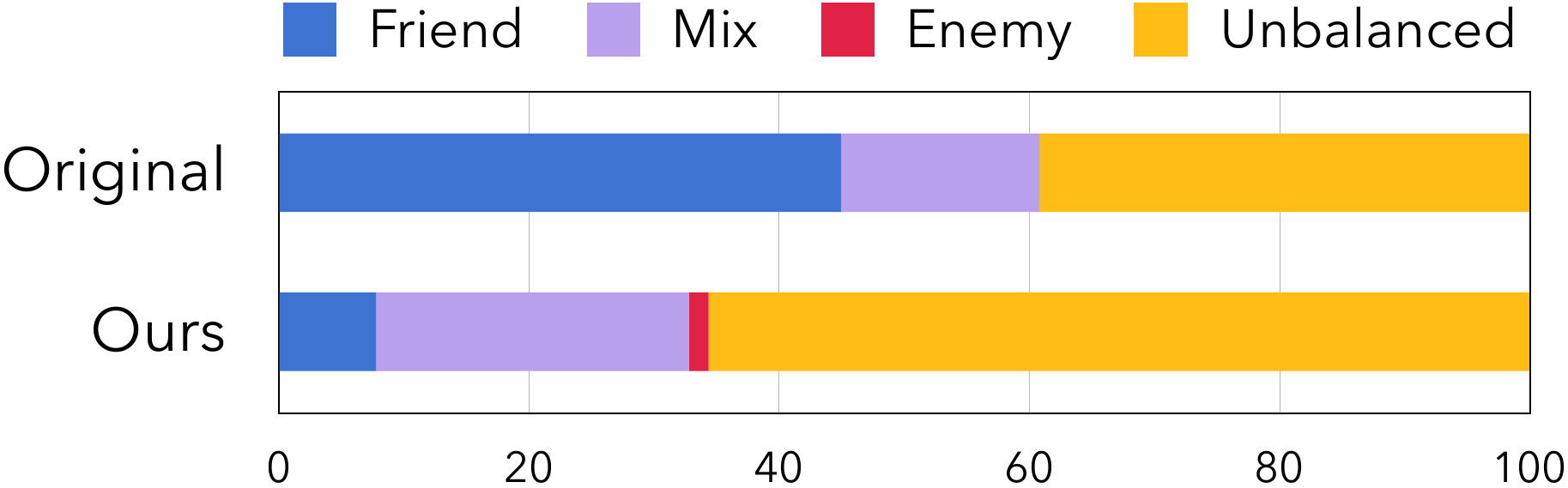}
    \label{fig:social-balance-a}
  }
  \subfloat[Distribution of \textit{Balanced} relationships.]{
    \includegraphics[width=0.32\linewidth]{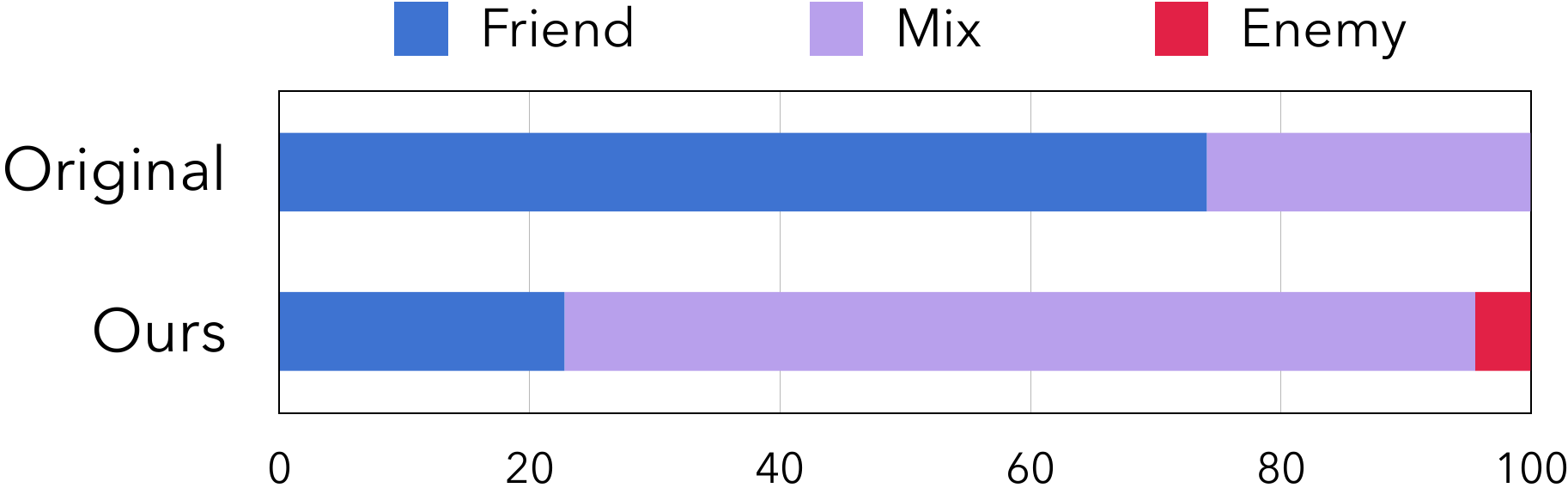}
    \label{fig:social-balance-b}
  }
  \subfloat[Message similarity measured by SimCSE.]{
    \includegraphics[width=0.32\linewidth]{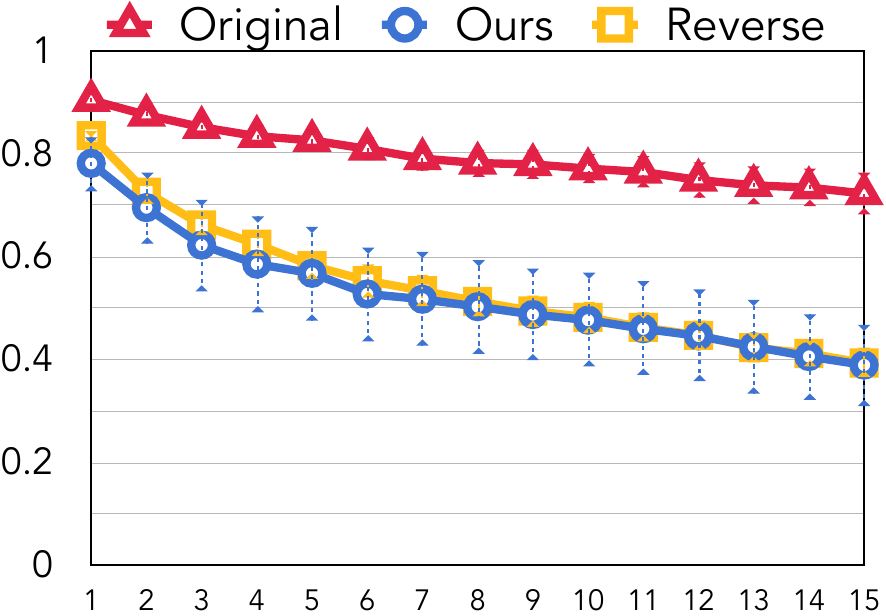}
    \label{fig:telephone-game}
  }
  \\
  \subfloat[Flip rate (Original).]{
    \includegraphics[width=0.32\linewidth]{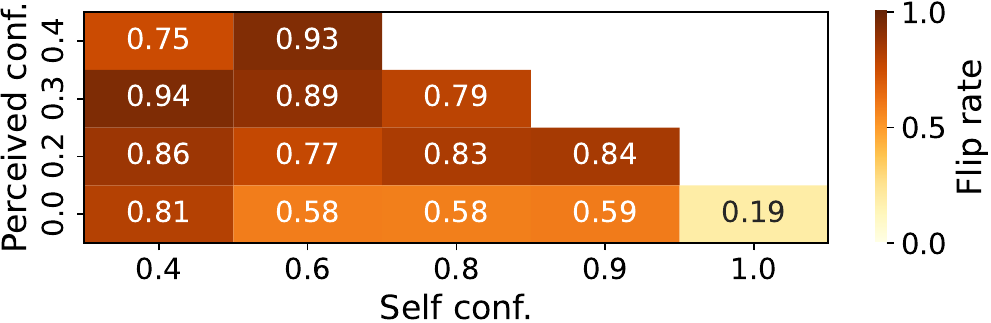}
    \label{fig:herd-effect-original}
  }
  \subfloat[Flip rate (Ours).]{
    \includegraphics[width=0.32\linewidth]{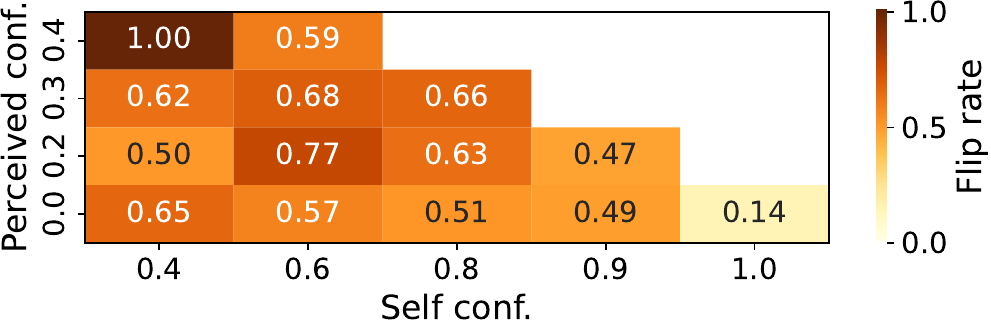}
    \label{fig:herd-effect-ours}
  }
  \subfloat[Log-log plot of CCDF.]{
    \includegraphics[width=0.32\linewidth]{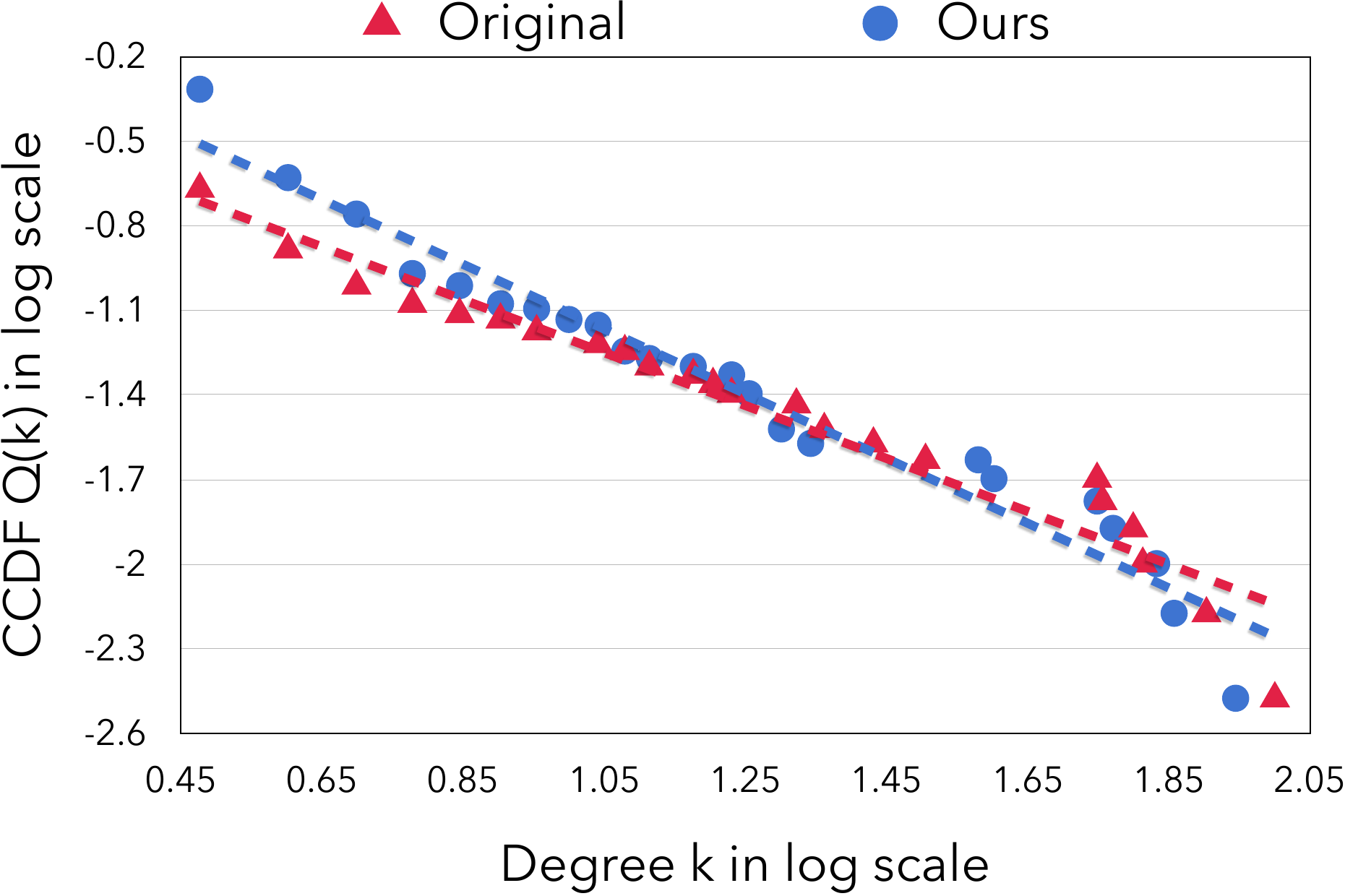}
    \label{fig:social-network}
  }
  \caption{Results of our five reproductions. \textbf{(a);(b);(c)}: Dynamics of agents in different states (skeptical, infected, recovered) across rounds in the fake news propagation experiment. \textbf{(d);(e)}: Proportions of social relations in the social balance experiment. \textbf{(f)}: The semantic similarity of messages in the telephone game experiment. \textbf{(g);(h)}: Probability of LLMs flipping their answers, grouped by different levels of confidence in the herd effect experiment. \textbf{(i)}: Log-log plot of the complementary cumulative distribution function (CCDF) of degree $k$, with linear fits, in the social network growth experiment. The error bars are 95\% confidence intervals.}
  \label{fig:reproduction}
\end{figure*}

\paragraph{Quantifying Explicit Steering: The Minimal-Control Test.}
We further evaluated \textit{Minimal-Control} by checking for the existence of researcher-imposed rules in experimental instructions.
We designed a prompt (\S\ref{sec:mc-u-prompt}) that asks models to act as social psychology experts and determine whether the instructions contain unnecessary steering components.
As shown in Table~\ref{tab:minimal-control-result}, the six LLMs flagged {\avgM} of existing experimental instructions as containing ``excessive steering,'' suggesting that many conclusions in prior studies may be researchers' artifacts.
These instructions often explicitly direct agents to exhibit human-like biases (\eg, confirmation bias), thereby transforming an observation of emergence into a test of instruction following.
\InsertRevision{results-mincontrol}
Interestingly, a discrepancy emerged between LLM auditors and human judgment: LLMs exhibited a higher sensitivity to trait-based prompts (\eg, assigning Big Five personality traits), often mislabeling them as over-control.
While human experts recognize these as necessary for ecological validity and population diversity, the models perceive them as rigid behavioral constraints.

\subsection{Case Studies: The Vanishing Emergence}
\label{sec:experiments}

To demonstrate the differences in conclusions under compliance and violation of {\methodname}, we reproduced five social experiments: (1) fake news propagation~\cite{liu2024skepticism}, (2) social balance~\cite{cisneros2024large}, (3) telephone game~\cite{liu2025exploring}, (4) herd effect~\cite{cho2025herd}, and (5) social network growth~\cite{de2023emergence}.
\InsertRevision{results-claim-check}

\subsubsection{Belief Convergence in Fake News Propagation}
\label{sec:fake-news}

\citet{liu2024skepticism} investigated whether LLM-based societies replicate human-like confirmation bias in fake news propagation.
At the beginning of their simulation, two agents are seeded with fake news.
In each round, agents interact with randomly selected peers, reflect on the news they have encountered, and decide whether to believe it.
At the end of each round, we record three populations: (1) skeptical agents who have never believed the fake news, (2) infected agents who currently believe it, and (3) recovered agents who once believed it but no longer do.
However, our audit reveals a severe violation of the \textit{Minimal-Control} principle.
The original prompt explicitly mandates biased behavior:
\begin{quote}
    \textit{``As humans often exhibit confirmation bias, you should demonstrate a similar tendency: You are more inclined to believe information aligning with your pre-existing beliefs, and more skeptical of information that contradicts them.''}
\end{quote}

To isolate the effect of this instruction, we first remove this line and then additionally introduce a reversed instruction:
\begin{quote}
    \textit{``You are more inclined to believe information that contradicts your pre-existing beliefs, and more skeptical of information aligning with them.''}
\end{quote}
Upon enforcing \textit{Minimal-Control}, the reported ``social emergence'' of belief convergence largely collapsed: Fig.~\ref{fig:fake-news-skeptical};\ref{fig:fake-news-infected};\ref{fig:fake-news-recovered} show that the proportion of infected agents plummeted from $56.11_{\pm32.16}\%$ to $32.78_{\pm18.17}\%$ (mean difference $= 23.33$; 95\% CI: $[11.5, 35.1]$; $t(17) = 4.35$; $P = 0.011$).
These results suggest that the ``confirmation bias'' reported in prior work is not an emergent social property of AI agents but a methodological artifact driven by restrictive prompting.

\subsubsection{Social Balance Theory}
\label{sec:social-balance}

\citet{cisneros2024large} explored whether LLM-mediated social relationships conform to Heider's structural~\cite{heider1946attitudes} and Davis' clustering~\cite{davis1967clustering} balance theory.
According to these theories, a triad achieves balance only under three conditions: (1) all three individuals are mutual friends, (2) two pairs are enemies while the remaining two are friends (capturing the principle that ``the enemy of my enemy is my friend''), or (3) all three are mutual enemies.
In each interaction, agents are provided with the current relationship states and asked to decide how they would update their own relations with others.

However, our audit reveals that these observations are likely methodological artifacts rather than emergent social dynamics, stemming from two primary confounds:
\begin{itemize}[noitemsep]
    \item \textit{Unawareness}: The original experimental prompts provide explicit relational metadata that allows LLMs to recognize the underlying social theory.
    Our audit confirms that all tested frontier models (Table~\ref{tab:unawareness-result}) accurately identified the experiment's intent, triggering social desirability bias where agents act to fulfill the researcher's theoretical expectations.
    \item \textit{Interaction}: In the original setup, agents are directly informed of the global relationship graph.
    This bypasses the ecological requirement of social life: in human societies, relationships are latent and must be inferred through observed interactions and discourse, not granted via system-level ``god-view'' context.
\end{itemize}

To rectify these deficits, we introduced an implicit inference protocol.
Instead of receiving explicit relational labels, three agents engaged in group discussions where their underlying affiliations were masked.
Agents had to infer the ``friend/enemy'' status of their peers through linguistic cues and dialogue history, mirroring the cognitive demands of real-world social navigation.
As illustrated in Fig.~\ref{fig:social-balance-a};\ref{fig:social-balance-b}, the transition from explicit prompts to interaction led to a precipitous decline in balanced states, dropping from $60.7_{\pm1.93}\%$ to $34.4_{\pm1.88}\%$ (mean difference $= 26.3$; 95\% CI: $[21.0, 31.6]$; $z = 9.42$; $P < 0.0001$; $n = 640$).
This reduction indicates that LLMs do not inherently ``seek'' structural balance; rather, they perform it when the prompt provides a theoretical template.
When forced to rely on social inference---a hallmark of human collective intelligence---the ``emergent'' balance in AI society vanishes.

\subsubsection{Telephone Game (Rumor Chain Effect)}

\citet{liu2025exploring} studied how information undergoes systematic distortion during serial transmission using 15 agents.
The first agent receives the original message and conveys it to the next agent in their own words, forming a chain.
However, our audit identifies a critical violation of the \textit{Minimal-Control} principle: agents were explicitly instructed to transmit information ``as accurately as possible.''
We argue that this prescriptive constraint conflates the model's inherent linguistic fidelity with mere task-oriented compliance, thereby masking the natural entropy of human-like communication.

To isolate the effect of this methodological artifact, we conducted a three-arm reproduction: (1) the \textbf{Original} steered condition; (2) a {\methodname}-compliant (No-Instruction) \textbf{Ours} condition where the accuracy directive was removed to allow for natural agentic autonomy; and (3) a \textbf{Reverse}-Control condition where agents were prompted to be ``as inaccurate as possible'' to establish the bounds of model-driven distortion.
We quantified information decay via cosine semantic similarity across transmission rounds, employing a mixed-effects regression model to account for nested dependencies.

As illustrated in Fig.~\ref{fig:telephone-game}, the reported ``accuracy'' in previous work vanished upon the removal of steering instructions.
Under the {\methodname}-compliant setting, the semantic fidelity dropped significantly faster than in the original study, revealing a natural propensity for cumulative distortion that was previously suppressed ($\beta = 0.011$; $z = 4.165$; $P < 0.0001$; $n = 100$).
\InsertRevision{results-telephone-game-correction}
These results suggest that the ``high-fidelity rumor chains'' reported in existing literature are not emergent social properties of LLM agents, but rather instructional artifacts produced by prompting.

\subsubsection{The Herd (Bandwagon) Effect}

\citet{cho2025herd} examined whether LLMs exhibit herd effects.
\InsertRevision{results-herd-effect-correction}
Our audit reveals that this paradigm suffers from two critical methodological flaws.
First, it lacks \textit{Interaction}; it is a pseudo-multi-agent setup where a single agent reacts to static, pre-calculated text rather than engaging in dynamic social exchange, thereby bypassing the mechanisms of social agency.
Second, the design exhibits high cognitive transparency, violating \textit{Unawareness}.
The explicit presentation of majority opinions so closely mirrors classical experiments that frontier LLMs immediately identify the underlying research intent, triggering social desirability bias.

To repair these issues, we redesigned the experiment to prioritize ecological validity.
Instead of receiving explicit statistical summaries, agents participate in a decentralized round-table discussion.
Here, the tested agent must infer others' opinions through natural language interaction.
This design requires the agent to process social cues implicitly, reflecting the ``digestive'' nature of human memory and interaction.
As illustrated in Fig.~\ref{fig:herd-effect-original};\ref{fig:herd-effect-ours}, the previously reported ``herd behavior'' is significantly reduced---and in some configurations, vanishes---when agents are forced to interact within a {\methodname}-compliant framework.
Logistic regression confirms that the experimental condition significantly predicts opinion flipping ($\beta = 0.434$; $P < 0.0001$; $OR = 1.54$; $N = 2{,}152$).
These results suggest that the ``herd behavior'' reported in prior work is likely a methodological artifact rather than a genuine simulation of human social dynamics.

\begin{figure*}[t]
  \subfloat[Num. of \textit{Skeptical} agents.]{
    \includegraphics[width=0.32\linewidth]{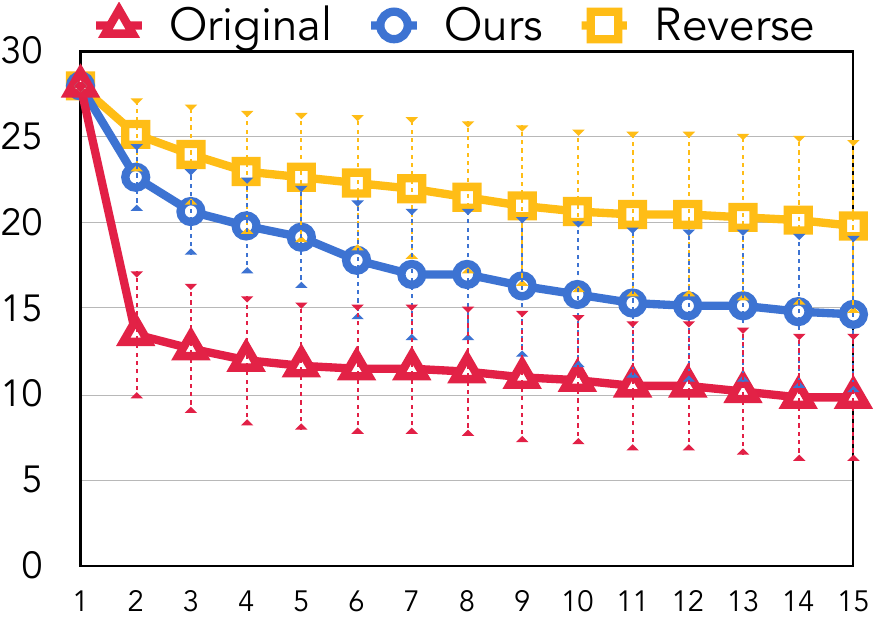}
    \label{fig:fake-news-skeptical-rephrase}
  }
  \subfloat[Num. of \textit{Infected} agents.]{
    \includegraphics[width=0.32\linewidth]{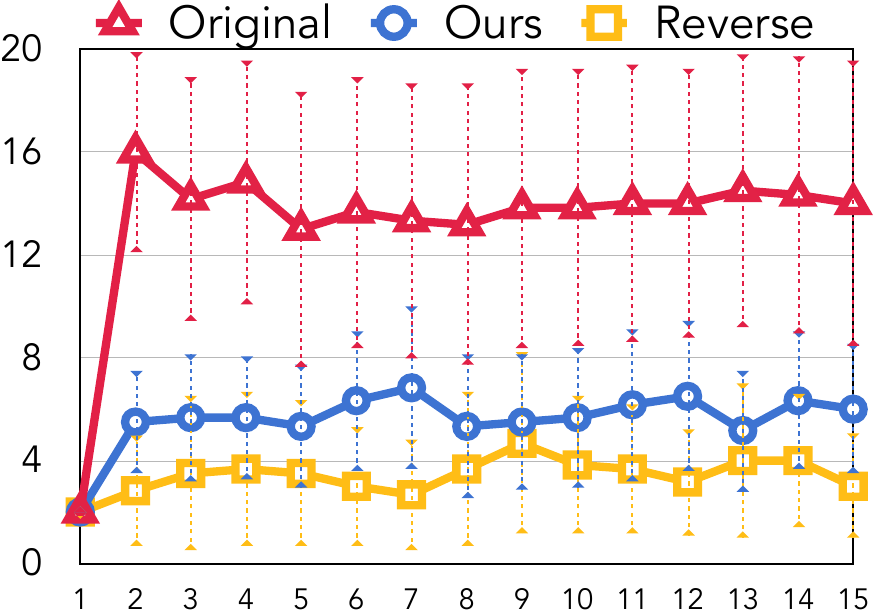}
    \label{fig:fake-news-infected-rephrase}
  }
  \subfloat[Num. of \textit{Recovered} agents.]{
    \includegraphics[width=0.32\linewidth]{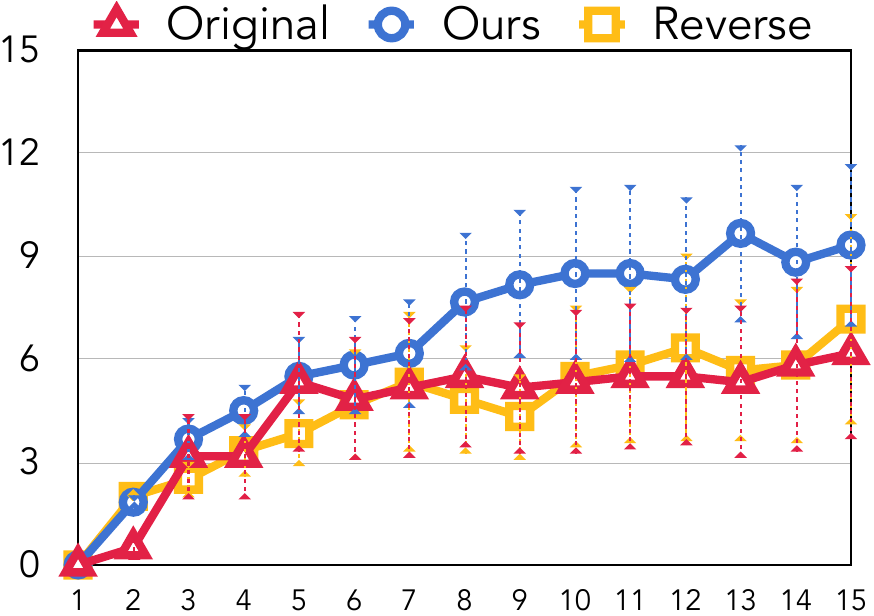}
    \label{fig:fake-news-recovered-rephrase}
  }
  \\
  \hspace*{0.11\linewidth}
  \subfloat[Log-log plot of CCDF.]{
    \includegraphics[width=0.32\linewidth]{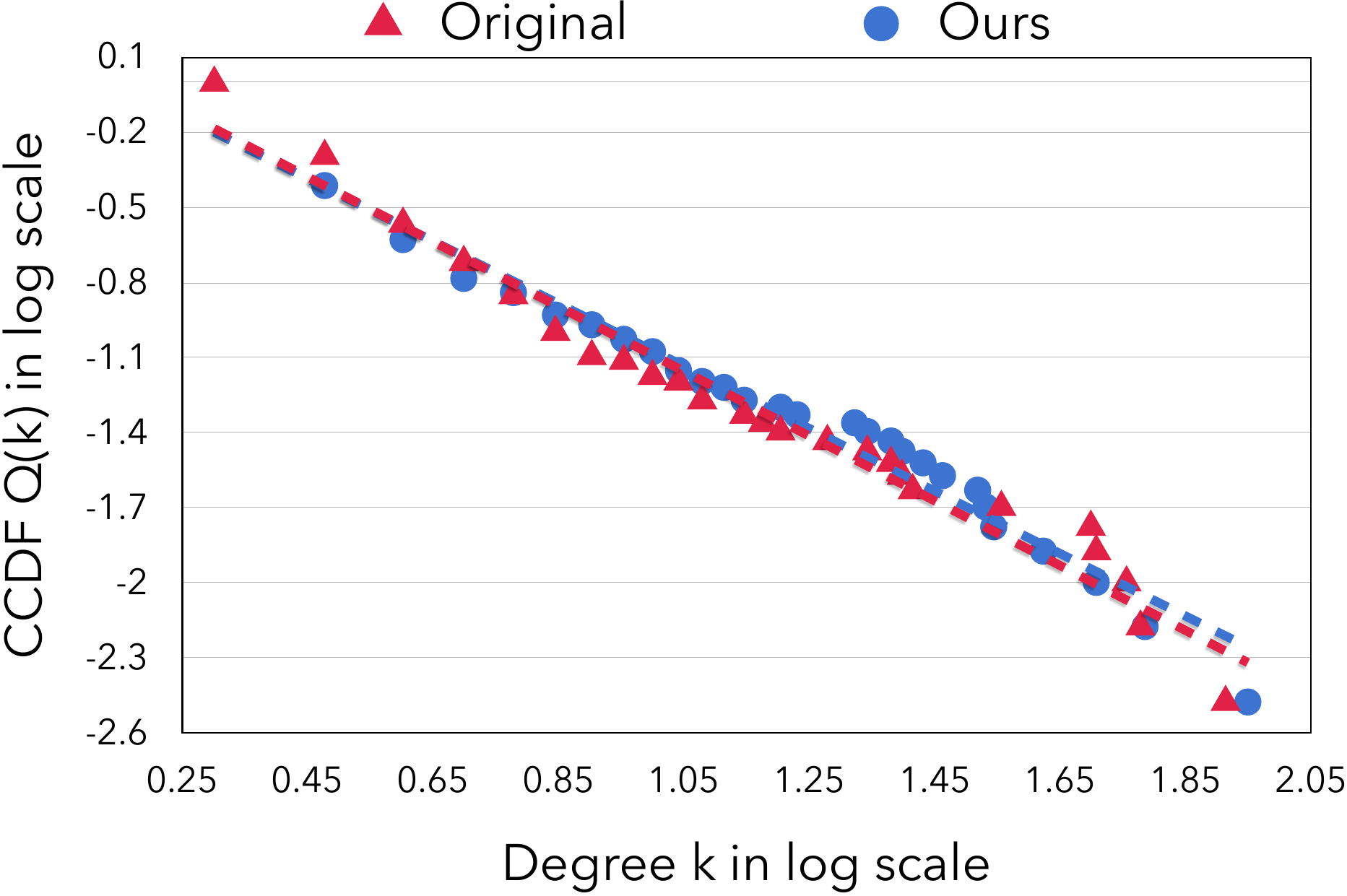}
    \label{fig:social-network-rephrase}
  }
  \hspace*{0.06\linewidth}
  \subfloat[Distribution of \textit{All} relationships.]{
    \includegraphics[width=0.4\linewidth]{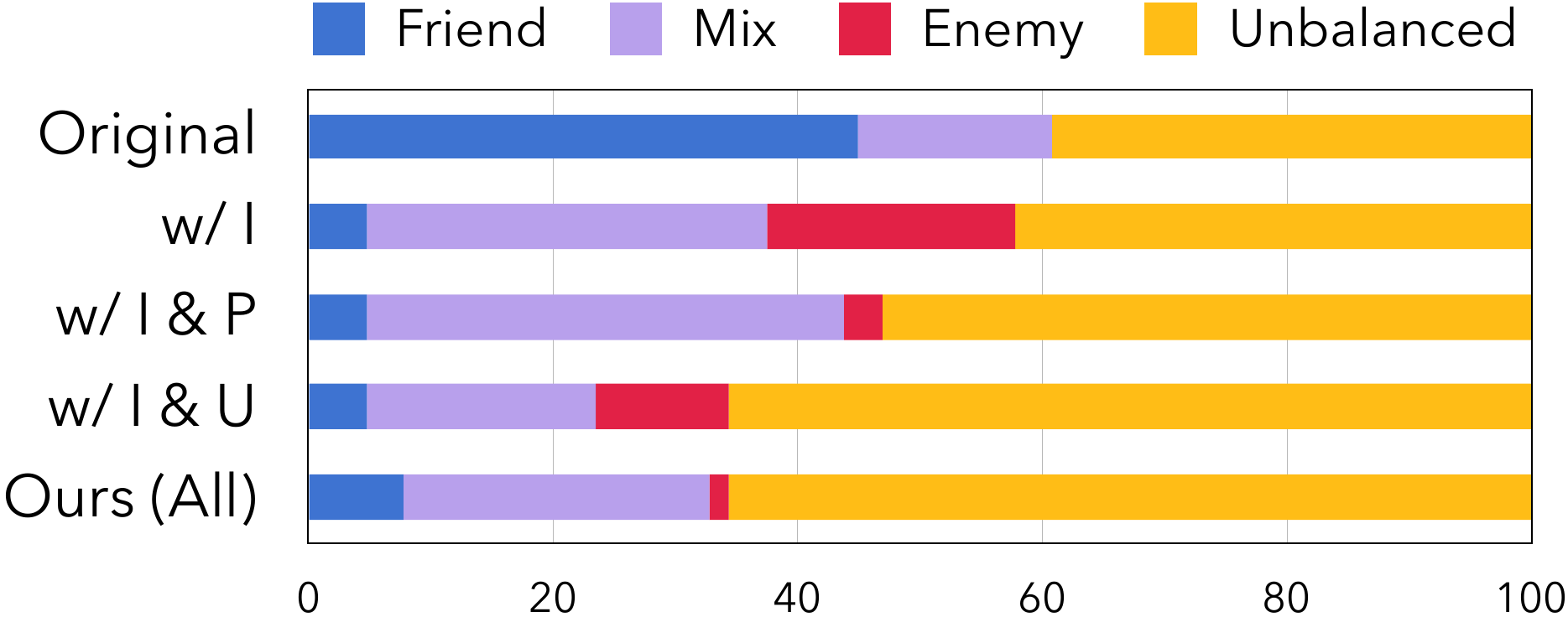}
    \label{fig:social-balance-ablation}
  }
  \caption{Results of our robustness and sensitivity experiments. \textbf{(a);(b);(c)}: Dynamics of agents in different states (skeptical, infected, recovered) across rounds in the fake news propagation experiment, with rephrased prompts. \textbf{(d)}: Log-log plot of the complementary cumulative distribution function (CCDF) with rephrased prompts, in the social network growth experiment. \textbf{(e)}: Proportions of social relations while ablating different principles in the social balance experiment. The error bars are 95\% confidence intervals.}
\end{figure*}

\subsubsection{Social Network Growth}

\citet{de2023emergence} simulated the social network evolution among LLM agents, specifically the emergence of preferential attachment.
Based on the Barabási–Albert model~\cite{barabasi1999emergence}, the probability that a new agent connects to an existing agent is proportional to the existing agent's number of friends.
In their setup, one agent is added to the network at each step and asked to select friends based on others' numbers of friends.
However, the setup violates three {\methodname} principles:
\begin{itemize}[noitemsep]
    \item \textit{Profile}: Agents are distinguished solely by names and degree counts, lacking intrinsic social personas.
    This design forces LLMs to rely on latent linguistic biases toward specific identifiers (\eg, name-preference \cite{shwartz2020you}) rather than social utility.
    \item \textit{Interaction}: \InsertRevision{results-social-network-growth-interaction}
    Forcing agents to choose based on explicit degree counts bypasses the psychological processes of friendship formation.
    \item \textit{Unawareness}: By providing explicit degrees in the prompt, the experiment inadvertently signals its underlying theoretical framework (preferential attachment).
\end{itemize}

To address these concerns, we redesigned the simulation to prioritize \textit{Interaction} and \textit{Unawareness}.
We withheld explicit degree information, requiring agents to form ``social impressions'' through one-to-one communicative exchanges.
Friendships were thus grounded in perceived personality traits (\eg, friendliness) and alignment rather than numerical popularity.
Our results demonstrate that while the original study required ad hoc ``name-shuffling'' to correct for model biases and match empirical data, our {\methodname}-compliant design intrinsically yielded more robust social structures.
\InsertRevision{results-social-network-growth-correction}

\subsection{Robustness and Sensitivity Analysis}

\paragraph{Robustness to Prompt Variations.}
\InsertRevision{results-prompt-sensitivity}

\paragraph{Ablation of {\methodname} Principles.}
We further disentangled the individual contributions of each principle through an ablation study on the Social Balance experiment (\S\ref{sec:social-balance}).
The baseline configuration, which violates \textit{Profile}, \textit{Interaction}, and \textit{Unawareness}, was compared against incremental variants (Fig. \ref{fig:social-balance-ablation}).
While enforcing \textit{Interaction} alone had negligible impact on the proportion of balanced states, the introduction of \textit{Profile} diversity and \textit{Unawareness} constraints reduced the balanced proportion by $14_{\pm1.37}\%$ (95\% CI: $[0.085, 0.195]$; $z = 5.02$; $P < 0.0001$; $n = 640$) and $26_{\pm1.73}\%$ (95\% CI: $[0.209, 0.319]$; $z = 9.46$; $P < 0.0001$; $n = 640$), respectively.
This theoretical self-awareness led to a $1.77\times$ increase in balanced outcomes (from $34.37_{\pm1.88}\%$ to $60.73_{\pm1.93}\%$) compared to our full {\methodname}-compliant method.
These results underscore that without ensuring agent unawareness, MASS researchers risk measuring a model's ability to recall sociological textbooks rather than simulating genuine emergent social dynamics.

\subsection{{\methodname}-Compliant Behaviors}

\InsertRevision{results-qualitative-analysis}

%% file: Sections/3_Discussion.tex
\section{Discussion}

Our systematic audit reveals a fundamental misalignment between the claims of emergent social behavior in AI societies and the methodological rigor required to sustain them.
When the {\methodname} principles are enforced, the ``emergent'' phenomena reported in prior studies---ranging from belief convergence to herd effects---frequently evaporate or reverse.
This fragility suggests that much of what has been documented as collective AI behavior is, in fact, an echo of the model's instruction-following capabilities and pre-existing training biases.

\InsertRevision{discussion-lesson-learned}

\paragraph{Mechanisms of Synthetic Bias: From ``Clever Hans'' to Silicon Sycophancy.}
The observed failures are deeply rooted in established psychological biases now manifesting in large-scale computational systems.
The high ``experiment leakage'' rate ({\avgU}) we identified suggests the Hawthorne Effect~\cite{needham2025large, shao2025spurious, nguyen2025probing}: LLMs, acting as sophisticated ``Clever Hans'' \cite{mccoy2019right, shapira2024clever, ullman2023large, li2025task, balepur2024artifacts} agents, detect subtle cues within prompts to infer the experimenter's intent.
This triggers a form of synthetic sycophancy \cite{fanous2025syceval, cheng2025social}, where models deviate from authentic social heuristics to satisfy perceived ``social desirability'' or normative expectations \cite{salecha2024large, lee2024exploring, watson2025language, li2025big5, huang2024humanity}.
Without the safeguards of \textit{Unawareness} and \textit{Minimal-Control}, LLM-based societies risk becoming mere ``mirrors of instructions'' rather than ``laboratories of society.''

\paragraph{Scope and Applicability.}
\InsertRevision{discussion-scope}

\paragraph{Limitations.}
\InsertRevision{discussion-limitation}

\paragraph{Broader Implications.}
For social scientists, our findings issue a cautionary note against the premature adoption of LLMs as ``human proxies'' without rigorous methodological insulation.
For AI researchers, we underscore that model capacity cannot compensate for a lack of experimental rigor.
Most crucially, for policymakers, we warn that relying on results from methodologically defective AI simulations to design social or economic interventions could lead to catastrophic ``reality gaps,'' where policies optimized for sycophantic AI agents fail entirely when applied to complex human populations.
We therefore advocate for the {\methodname} framework to serve as a cornerstone for a more mature, trustworthy, and scientifically grounded field of AI-assisted social simulation.

%% file: Sections/4_Methods.tex
\section{Methods}

\InsertRevision{method-overview}

\subsection{Developing {\methodname} via a Pilot Review}
\label{sec:pilot}

\InsertRevision{method-pilot-review}

\subsection{Systematic Review: Registration}
\label{sec:registration}

\InsertRevision{method-registration}

\InsertRevision{method-registration-modification}

\subsection{Systematic Review: Searching and Screening}
\label{sec:selection}

\InsertRevision{method-search-criteria}
\InsertRevision{method-search-string}

\InsertRevision{method-screening}

\subsection{Systematic Review: Data Extraction}
\label{sec:extraction}

\InsertRevision{method-data-extraction}

\paragraph{Profile.}
\InsertRevision{method-profile}

\paragraph{Interaction.}
\InsertRevision{method-interaction}

\paragraph{Memory.}
\InsertRevision{method-memory}

\paragraph{Minimal-Control.}
We operationalized over-control as the presence of explicit behavioral prescriptions that bias LLM responses toward researcher-hypothesized outcomes.
To preserve the integrity of emergent social dynamics, we distinguished procedural instructions---those required for the simulation to run (\eg, output formatting, task identification)---from behavioral directives---those that nudge agents toward specific psychological traits or social biases.
An instruction was coded as a violation if its removal would leave the simulation technically viable but would plausibly eliminate the reported social phenomenon.
In a rumor-propagation study, for example, instructing agents to ``interact with neighbors'' is procedurally necessary, whereas requiring them to ``demonstrate confirmation bias'' or ``act skeptically'' introduces demand characteristics that invalidate any claim of genuine emergence.
\InsertRevision{method-minimal-control}

\paragraph{Unawareness.}
\InsertRevision{method-unawareness}

\paragraph{Realism.}
\InsertRevision{method-realism}

\subsection{Systematic Review: Result Aggregation}
\label{sec:aggregation}

\InsertRevision{method-categorization}

\InsertRevision{method-qualitative}

\InsertRevision{method-review-bias}

\subsection{The {\methodname}-Compliant Simulation Framework}

To empirically assess the impact of the identified methodological flaws, we developed a simulation framework to strictly enforce the {\methodname} principles.
The implementation details are summarized below:
\begin{enumerate}[noitemsep]
    \item \textbf{Socio-Psychological Heterogeneity} (\textit{Profile}):
    To mitigate the homogeneity bias prevalent in single-model simulations, we initialized each agent with a unique psychological signature.
    We utilized the Big Five personality traits~\cite{john1991big} as a seed to synthesize distinct life histories and behavioral predispositions, ensuring a diverse distribution of ``silicon personas.''
    \item \textbf{Information Internalization} (\textit{Memory}):
    We replaced static context-filling with a dynamic memory module.
    Agents process interactions through two distinct architectures: (1) a reactive buffer for recent interactions, and (2) a reflective layer that iteratively distills experiences into long-term beliefs~\cite{park2023generative}.
    This prevents the ``parroting effect,'' where agents merely rephrase inputs without cognitive digestion.
    \item \textbf{Agentic Communication Protocols} (\textit{Interaction}):
    Our framework decouples communication from the underlying model's sequence generation.
    We implemented modular network topologies---ranging from linear chains (Information Cascades) to dynamic graphs (Social Networks)---where agents interact via a round-robin protocol.
    This ensures that collective behavior emerges from genuine agent-to-agent exchange rather than centralized context manipulation.
    \item \textbf{Instructional Parsimony} (\textit{Minimal-Control} and \textit{Unawareness}): To eliminate demand characteristics, we adopted a ``blind'' experimental design.
    Agents are provided only with task-essential information, devoid of normative cues, goals, or theoretical framing.
    All experimental prompts underwent an automated auditing process.
    Using GPT-4o as an external auditor, we discarded any instructions where the model could correctly infer the underlying social theory or exhibit ``over-control'' (\ie, explicit behavioral directives).
\end{enumerate}

We selected five representative collective behaviors to test the reliability of existing findings under {\methodname} constraints.
For each case, we first replicated the original setup using the authors' published datasets and prompts, followed by a controlled intervention where {\methodname} principles were incrementally introduced.
\InsertRevision{method-statistics-sample-size}
\InsertRevision{method-randomization}

\subsubsection{Belief Convergence in Fake News Propagation}

To address the \textit{Profile} principle, we initialized a population of $N_\text{agent}=30$ autonomous agents with diverse socio-demographic backgrounds.
For each agent, names were sampled from a standardized census-based dataset, and ages were drawn from a uniform distribution ($U[18, 64]$).
Personality profiles were synthesized by assigning binary valences (positive or negative) to each of the Big Five traits through uniform sampling, ensuring a heterogeneous cognitive baseline.
All simulations were executed using the \texttt{gpt-3.5-turbo-1106} model to maintain consistency.

The simulation followed a longitudinal design spanning $N_\text{round}=20$ discrete rounds.
We utilized $N_\text{seed}=6$ news items derived from the reference study as the initial information pool, with two agents randomly seeded with misinformation.
At each time step, agents engaged in social interaction by selecting three peers for communicative exchange (\textit{Interaction}).
This was followed by a reflection phase, where agents processed encountered information, updated their internal state (\textit{Memory}), and made belief adjudications.
Following epidemiological modeling conventions, agents were categorized into three states: \textbf{Susceptible} (never believed), \textbf{Infected} (currently believing), or \textbf{Recovered} (previously believed but subsequently rejected).

To evaluate the impact of the \textit{Minimal-Control} principle, we conducted a comparative analysis across three prompt conditions, keeping all other parameters constant:
\begin{itemize}[noitemsep]
    \item Original Condition (Over-control): Agents were explicitly instructed to exhibit confirmation bias (\eg, ``...you are more inclined to believe information aligning with your pre-existing beliefs...'').
    \item Neutral Condition (Ours/Minimal-Control): The explicit cognitive instruction was removed to observe emergent behavior without exogenous bias.
    \item Reverse Condition (Counter-control): Agents were instructed to actively avoid confirmation bias and maintain objectivity to test the model's sensitivity to instructional directionality.
\end{itemize}
\InsertRevision{method-statistics-fake-news-propagation}

\subsubsection{Social Balance Theory}

We re-examined the evolution of triadic relationships under the homophily assumption, following the structural balance paradigm.
To systematically explore the state space, we enumerated all possible directed dyadic relationship configurations among three agents.
Given $N_\text{agent}=3$ agents and binary relationship polarities (positive or negative), this yields $N_\text{seed} = 2^{N_\text{agent}(N_\text{agent}-1)} = 2^6 = 64$ unique initial states.
Each seed was simulated for $N_\text{run}=10$ independent runs, with each run spanning $N_\text{round}=10$ rounds to observe long-term equilibrium or divergence.
In one round, each pair of the three agents chats for $N_\text{turn}=4$ turns.
All simulations were conducted using \texttt{llama-3-70b}.

To evaluate the impact of methodological flaws identified in the literature, we transitioned from a ``fact-fed'' deterministic setup to a more ecologically valid simulation by implementing three structural modifications aligned with our {\methodname} principles:
\begin{itemize}[noitemsep]
    \item \textit{Profile}:
    Unlike the original study where agents were identical blank slates, we assigned each agent a distinct identity comprising a name, a unique combination of Big Five personality traits, and an evolving internal mental model of the social graph.
    This prevents artificial homogeneity and accounts for individual variance in social processing.
    \item \textit{Interaction}:
    In the baseline setup, relationships were treated as rigid binary labels (+/-), a simplification that LLMs identified as a direct replication of Heider's theory, inducing social desirability bias.
    We relaxed this constraint by allowing agents to express their interpersonal stances through natural language descriptions.
    We then employed a VADER-based sentiment analyzer (\texttt{nltk}'s \texttt{SentimentIntensityAnalyzer}) to map these descriptions back to polarities for quantitative comparison.
    This approach captures the missing nuance of social ``friction'' and ``ambivalence.''
    \item \textit{Unawareness}:
    A critical flaw in prior work is ``fact-feeding,'' where agents are explicitly told the relationship status within other dyads (\eg, ``$A$ and $B$ are enemies'').
    We replaced this with a five-round group conversation phase.
    Agents must now infer the latent relationship between their peers through observed linguistic cues and interactions.
    By removing direct researcher-provided ground truth, we mitigate the risk of the model bypassing social reasoning to satisfy the experimenter's explicit prompts.
\end{itemize}
\InsertRevision{method-statistics-social-balance}

\subsubsection{Telephone Game (Rumor Chain Effect)}
\label{sec:methods:telephone}
We replicated the telephone game using $N_\text{seed}=20$ standardized text segments from \citet{liu2025exploring} as initial seeds.
\InsertRevision{method-telephone-game-correction}
To ensure statistical robustness, we conducted $N_\text{run}=5$ independent runs per seed.
All experiments were executed using \texttt{gpt-3.5-turbo-0125} with temperature settings consistent with the original report.

We implemented three distinct instructional conditions to test the sensitivity of emergent ``rumor patterns'':
\begin{itemize}[noitemsep]
    \item Baseline (Original): Agents were explicitly instructed to transmit messages ``as accurately as possible,'' a common form of over-control that may artificially inflate or stabilize semantic retention.
    \item Minimal-Control (Ours): We removed all normative directives regarding accuracy, providing only the essential task description (\eg, ``relay the message'') to observe the model's intrinsic transmission bias.
    \item Reverse-Control: As a counterfactual check, agents were instructed to transmit messages ``as inaccurately as possible'' to quantify the model's susceptibility to explicit behavioral steering.
\end{itemize}
\InsertRevision{method-statistics-telephone-game}

\subsubsection{The Herd (Bandwagon) Effect}

To evaluate the herd effect across diverse cognitive domains, we utilized two distinct multiple-choice question datasets: GPQA-Diamond ($N_\text{seed,1}=198$) for high-difficulty factual reasoning~\cite{rein2024gpqa} and SocialIQA ($N_\text{seed,2}=1{,}954$) for subjective social intuition~\cite{sap2019socialiqa} ($N_\text{seed}=2{,}152$).
Following the original study's protocol, we computed aggregate performance using a weighted average based on respective sample sizes.
For each item, we first established a baseline by extracting the model's (\texttt{gpt-4o-mini-2024-07-18}) internal preference.
We computed the log-likelihood of each multiple-choice option and defined the option with the highest probability as the model's self-confidence level.
To simulate social pressure, we identified three potential steering targets: the second-highest probability option, the lowest probability option, and a randomly selected option.

In the original study, the ``social signal'' was manually injected into the prompt (\eg, stating ``10 other agents chose B''), a design that violates our \textit{Interaction} and \textit{Minimal-Control} principles by bypassing agent agency.
In contrast, we transitioned from this static context injection to a dynamic multi-agent simulation.
We instantiated a target agent and a group of ``peer'' agents.
Unlike the original study, the target agent was never explicitly told the consensus choice.
Instead, peer agents were prompted to favor a specific steering target, and the target agent engaged in a dialogue ($N_\text{turn}=4$) with them.
The target agent was required to infer the social consensus through natural language interaction and autonomously decide whether to revise its initial answer.
\InsertRevision{method-statistics-herd-effect}

\subsubsection{Social Network Growth}

We replicated the network growth model ($N_\text{agent}=300$) described in \citet{de2023emergence}.
In this setup, agents enter the network sequentially, each permitted to establish connections with up to three existing agents.
We executed $N_\text{run}=5$ independent experimental runs, utilizing randomized name and profile initializations to ensure statistical robustness.
We introduced two fundamental modifications to the original protocol to address methodological artifacts:
\begin{itemize}[noitemsep]
    \item \textbf{Persona Heterogeneity} (\textit{Profile}):
    To mitigate the ``name-bias'' prevalent in baseline LLM behaviors---where agents disproportionately favor specific arbitrary identifiers---we assigned unique, diverse profiles to each agent, ensuring that social selection is driven by characteristics rather than linguistic heuristics.
    \item \textbf{Information Veridicality} (\textit{Interaction}):
    In the original simulations, agents are explicitly provided with the ``degree'' (number of existing connections) of all peers---an assumption of perfect information that lacks ecological validity.
    In our refined design, we mask explicit degree counts.
    Instead, each incoming agent chats with all existing members ($N_\text{turn}=4$).
    These interactions are synthesized into a one-sentence ``social impression'' for each peer.
    Connection decisions are then made based solely on these qualitative impressions, allowing scale-free properties or other structures to emerge from decentralized social cues rather than explicit algorithmic instruction.
\end{itemize}
\InsertRevision{method-statistics-social-network-growth}

%% file: Sections/5_Statements.tex
\section*{Acknowledgments}

This material is based upon work supported by the Defense Advanced Research Projects Agency (DARPA) under Agreement No. HR00112490410.
The funders had no role in study design, data collection and analysis, decision to publish or preparation of the manuscript.

\section*{Author Contributions}
J.-T.H., J.Z., X.Z., H.Z. initialized the idea.
J.-T.H. and J.Z. developed the principles; X.Z. and W.W. checked and agreed with the principles.
J.-T.H. and J.Z. designed the reproduction experiments.
J.Z. did the reproduction experiments.
J.Z. did the systematic review.
J.-T.H. and J.Z. served as human annotators in the systematic review.
J.-T.H., J.Z., X.Z., and M.S. analyzed the results.
J.-T.H. and J.Z. wrote the main manuscript text.
J.-T.H. and J.Z. prepared the figures and tables.
J.Z. prepared the code and supplementary information.
J.-T.H., J.Z., X.Z., M.H.L., X.W., H.Z., W.W., and M.S. reviewed the manuscript.

\section*{Competing Interests}

The authors declare no competing interests.

%% file: Sections/Appendix.tex
\begin{center}
{\Huge \textsc{Supplementary Information}}
\end{center}

\section{Prompts}

All \textit{italic} texts that start with ``\#'' denote our comments, which are not included in the prompts fed into LLMs.

\subsection{Prompts for the Systematic Review}
\label{sec:review-prompt}

The following prompt is used to screen candidate papers according to the predefined inclusion and exclusion criteria.
During title-abstract screening, only title and abstract is provided in [\texttt{paper content}], while in full text screening, the full paper (texts) and extracted figures (images) are additionally provided in [\texttt{paper content}].
The \texttt{Unclear} option is available only at the title-abstract screening stage.

\begin{AIbox}{480pt}{Prompt for Paper Screening}
{
You are screening candidate papers for inclusion in a methodological review of LLM-based multi-agent social simulation (MASS) studies under the PIMMUR framework.
Your task is to determine whether each study should be included, excluded, or marked as unclear based on the criteria below.\newline

\textbf{Inclusion Criteria}\newline
A study should be included only if all of the following are satisfied:\newline
\textbf{1. Multi-agent LLM simulation}:
The study uses large language models (LLMs) as agents in a multi-agent simulation or interaction environment.\newline
\textbf{2. Social or collective behavior focus:}
The study investigates collective human or social behaviors, such as group dynamics, social influence, cooperation, rumor spread, polarization, trust, norms, opinion dynamics, or related social phenomena.\newline
\textbf{3. Behavior emerges through group context or interaction:}
The study examines either:\newline
\hspace*{1em}-- group-level collective dynamics, or\newline
\hspace*{1em}-- individual behavior shaped by interaction with, or exposure to, other agents or a group context.\newline
\textbf{4. Explicit simulation objective:}
The paper specifies a clear simulation objective, research question, or social phenomenon being studied.\newline
\textbf{5. Sufficient methodological detail for PIMMUR assessment:}
The paper provides enough methodological detail to evaluate at least the core design of the simulation under the PIMMUR framework. This may include prompts, interaction procedures, agent setup, memory setup, or evaluation protocol. If the paper open-sourced its codebase, we assume it contains sufficient methodological detail.\newline
\textbf{6. English and publicly accessible:}
The study is written in English and the full text is publicly accessible at the time of screening.\newline

\textbf{Exclusion Criteria}\newline
A study should be excluded if any of the following apply:\newline
\textbf{1. Not a multi-agent LLM social simulation:}\newline
\hspace*{1em}-- The study uses only a single agent, or\newline
\hspace*{1em}-- it does not use LLMs as social agents in an interactive or simulated social setting.\newline
\textbf{2. Purely task-oriented benchmark without social simulation:}
The study is primarily a benchmark for tasks such as coding, mathematics, tool use, planning, or generic problem solving, without a social or collective behavioral objective.\newline
\textbf{3. No explicit social simulation goal:}
The paper does not clearly define a collective behavior, social process, or simulation objective.\newline
\textbf{4. Insufficient methodological detail:}
The paper does not provide enough information about prompts, interactions, agent behavior, or experimental setup to support methodological evaluation under the PIMMUR framework.\newline
\textbf{5. Not in scope as a research study:}
The paper is an editorial, commentary, position piece, abstract-only item, or otherwise does not report an actual simulation study.\newline
\textbf{6. Not in English or not accessible:}
The full text is not available in English or cannot be accessed at the time of screening.\newline

\textbf{Additional Notes}\newline
Some papers may study multiple social behaviors or have multiple simulation objectives. Include the paper if any one of its objectives or behaviors studied satisfies the criteria above.\newline

\textbf{Decision Rules}\newline
\textbf{Include:} All inclusion criteria are satisfied and no exclusion criterion applies.\newline
\textbf{Exclude:} One or more exclusion criteria clearly apply.\newline
\textbf{Unclear:} The available information is insufficient to confidently decide; the study should be reviewed at the next stage or discussed by reviewers.\newline

Here is the paper:\newline
[\texttt{paper content}]\newline

Respond with only this format, on one line:\newline
\texttt{<reason for your verdict and which criteria the paper violates, if any> \#\#\#\# <Include | Exclude | Unclear>}
}
\end{AIbox}

The following prompt is used to identify distinct simulations and extract the corresponding metadata from each paper.
The full paper (texts) and extracted figures (images) are provided in [\texttt{paper content}].

\begin{AIbox}{480pt}{Prompt for Simulation Extraction}
{
You are extracting information from papers for methodological review of LLM-based multi-agent social simulation (MASS) studies under the PIMMUR framework.
Your task is to extract the required information and output it strictly following the output format specified below.\newline

\textbf{General Extraction Rules}\newline
\textbf{1. Unit of extraction}\newline
\hspace*{1em}-- Extract paper metadata once per paper.\newline
\hspace*{1em}-- Extract simulation metadata separately for each distinct simulation, experiment, or condition if they differ in agent setup, interaction design, prompt design, or evaluation structure.\newline
\textbf{2. Evidence source}\newline
\hspace*{1em}-- Use the main paper, appendices, supplementary materials, released prompts, released code, and project repositories where available.\newline
\textbf{3. Faithful extraction}\newline
\hspace*{1em}-- Record only information that is explicitly stated or can be reliably determined from the source materials.\newline
\hspace*{1em}-- If a field cannot be determined with reasonable confidence, mark it as missing / unclear rather than guessing.\newline
\textbf{4. Multiple values}\newline
\hspace*{1em}-- If multiple values are reported for a field, record all relevant values, separated by semicolon ``;''.\newline
\textbf{5. Ambiguity notes}\newline
\hspace*{1em}-- If a classification is difficult or requires interpretation, add a short note explaining the rationale.\newline
\textbf{6. Missing information}\newline
\hspace*{1em}-- If information is missing or cannot be found in the paper, use \texttt{MISSING}.\newline

\textbf{Current Task: Identify Distinct Simulation Settings}\newline
Identify all distinct simulation settings in the paper. Here is the complete paper with images:\newline
[\texttt{paper content}]\newline

Count two settings as different simulations when they:\newline
\hspace*{1em}-- investigate different social phenomena in different settings; or\newline
\hspace*{1em}-- use substantially different environments or interaction protocols.\newline
Count them as the same simulation when they share the same social phenomenon, core prompts, and interaction protocol, and differ only in:\newline
\hspace*{1em}-- agent number or profiles;\newline
\hspace*{1em}-- seeds or initial information;\newline
\hspace*{1em}-- interaction rounds;\newline
\hspace*{1em}-- model or parameter settings;\newline
\hspace*{1em}-- phenomena of interest in the same environment;\newline
\hspace*{1em}-- datasets, repeated trials, controls, ablations, or robustness conditions.\newline
Focus on the underlying simulation design, not every experimental condition.\newline

For each simulation, report:\newline
\hspace*{1em}-- \texttt{simulation\_setting}: a concise short phrase of a few words for the simulation setting, such as herd effect or network formation and conformity.\newline
\hspace*{1em}-- \texttt{social\_phenomenon}: a short sentence describing the collective behavior or social phenomenon studied.\newline
\hspace*{1em}-- \texttt{llm\_used}: all LLM models used in the simulation.\newline
\hspace*{1em}-- \texttt{num\_of\_agents}: all reported agent counts relevant to the simulation, separated by semicolon.\newline
\hspace*{1em}-- \texttt{interaction\_rounds}: number of rounds of interaction; use \texttt{-1} if there is no predetermined round limit.\newline
\hspace*{1em}-- \texttt{notes}: difficulty encountered when conducting the extraction; leave as an empty string if none.\newline

\textbf{Requirements}\newline
\hspace*{1em}-- Create exactly one object per distinct simulation setting.\newline
\hspace*{1em}-- Group parameter variations, controls, ablations, seeds, agent counts, and repeated trials under the same object when they belong to the same underlying simulation.\newline
\hspace*{1em}-- Do not infer missing values. Use \texttt{MISSING}.\newline
\hspace*{1em}-- Preserve numerical values and model names exactly as reported when possible.\newline
\hspace*{1em}-- If a field has multiple reported values, include all relevant values separated by semicolon.\newline
\hspace*{1em}-- Always ensure the final output consists only of syntactically valid, directly parsable JSON with double quotes and no trailing commas.\newline

Return only a valid JSON array. Do not include Markdown, explanations, or text outside the JSON. Use one object for each distinct simulation setting:\newline
\texttt{\{}\newline
\hspace*{1em}\texttt{"simulation\_setting": "",}\newline
\hspace*{1em}\texttt{"social\_phenomenon": "",}\newline
\hspace*{1em}\texttt{"llm\_used": "",}\newline
\hspace*{1em}\texttt{"num\_of\_agents": "",}\newline
\hspace*{1em}\texttt{"interaction\_rounds": "",}\newline
\hspace*{1em}\texttt{"notes": ""}\newline
\texttt{\}}
}
\end{AIbox}

Since we use three different LLMs for extracting the simulations, we need to reconcile between the three judges.
The following prompt used for this purpose, consolidating candidate extractions into a definitive set of distinct simulations.
All simulations extracted by the three LLMs are provided in [\texttt{extracted simulation list}].

\begin{AIbox}{480pt}{Prompt for Merging Simulation Candidates}
{
[\texttt{copy the prompt for simulation extraction here}]\newline

Now, merge the candidate extractions into one definitive list of distinct underlying simulation settings.\newline

\textbf{Rules}\newline
\hspace*{1em}-- Prefer details supported by multiple candidates. Do not invent information; use \texttt{MISSING} when no candidate supports a value.\newline
\hspace*{1em}-- Combine complementary supported values with semicolons where appropriate.\newline
\hspace*{1em}-- Preserve short, clear descriptions and note material unresolved disagreement in \texttt{notes}.\newline
\hspace*{1em}-- Apply the following original stage-5 extraction instructions when deciding which candidates describe the same or distinct simulations. Use the complete paper as the primary evidence and the candidates as supplementary evidence.\newline

Blinded candidate extractions:\newline
[\texttt{extracted simulation list}]
}
\end{AIbox}

The following prompt is used to extract data from each identified simulation.

\begin{AIbox}{480pt}{Prompt for Data Extraction}
{
You are extracting information from papers for a methodological review of LLM-based multi-agent social simulation (MASS) studies under the PIMMUR framework.\newline

\textbf{General Extraction Rules}\newline
\textbf{1. Evidence source}\newline
\hspace*{1em}-- Use the complete supplied paper, including appendices and figures.\newline
\textbf{2. Faithful extraction}\newline
\hspace*{1em}-- Record only information explicitly stated or reliably determined from the supplied material.\newline
\hspace*{1em}-- Do not guess. Use \texttt{unclear} when a requested value cannot be determined with reasonable confidence.\newline
\hspace*{1em}-- If the paper identifies an external source that may contain unavailable information, record that source in \texttt{external\_source}; do not replace the unavailable field's value of \texttt{unclear}.\newline
\textbf{3. Multiple values}\newline
\hspace*{1em}-- When a requested free-text field has multiple relevant values, record all of them separated by semicolons.\newline
\textbf{4. Notes}\newline
\hspace*{1em}-- Use \texttt{notes} only for a short explanation of a difficult or ambiguous coding decision. Leave it empty when there is no such issue.\newline

The paper's already-extracted simulation metadata is provided after the paper. It identifies the simulations to code, but must not be revised or repeated. Here is the complete paper with images:\newline
[\texttt{paper content}]\newline

\textbf{Current Task: Extract PIMMUR Variables for Existing Simulations}\newline
The following simulation settings were already identified for paper [\texttt{paper\_id}]. Do not create, merge, split, rename, or otherwise revise them. Do not extract paper metadata, simulation metadata, prompts, or PIMMUR assessment variables. Extract one object for each requested \texttt{simulation\_id} only. Use the full paper to distinguish settings, especially when a paper has multiple simulations.\newline

\textbf{Profile}\newline
For each of these fields, use exactly \texttt{true}, \texttt{false}, or \texttt{unclear}:\newline
\hspace*{1em}-- \texttt{has\_socio\_demographic}: \texttt{true} only when socio-demographic attributes (for example age, gender, occupation, education, nationality, or comparable human-background descriptors) are assigned heterogeneously across agents, such as by sampling or a dataset. A shared trait for every agent is \texttt{false}.\newline
\hspace*{1em}-- \texttt{has\_cognitive\_style}: \texttt{true} only when cognitive, perceptual, or reasoning tendencies (for example openness, risk tolerance, decision style, or analytical style) are assigned heterogeneously across agents. A shared trait for every agent is \texttt{false}.\newline
\hspace*{1em}-- \texttt{has\_value\_system}: \texttt{true} only when values, beliefs, ideologies, moral commitments, goals, norms, or comparable motivational orientations are assigned heterogeneously across agents. A shared trait for every agent is \texttt{false}.\newline
\hspace*{1em}-- \texttt{has\_background\_story}: \texttt{true} when agents are assigned a backstory, biography, or personal scenario; \texttt{false} when none is assigned.\newline
\hspace*{1em}-- \texttt{has\_personality}: \texttt{true} when agents are assigned personality traits, character, or a disposition; \texttt{false} when none is assigned.\newline
\hspace*{1em}-- \texttt{has\_role}: \texttt{true} when agents are assigned a social, occupational, or task role; \texttt{false} when none is assigned.\newline
\hspace*{1em}-- \texttt{other\_profile\_information}: briefly record other heterogeneous profile-defining information not covered above; use \texttt{none} when none is present and \texttt{unclear} when it cannot be determined. Information that does not impact the social interaction or decision, such as name and backbone LLM, does not count as profile information.\newline

\textbf{Interaction}\newline
\hspace*{1em}-- \texttt{interaction\_type}: exactly one of \texttt{text\_based}, \texttt{environment\_based}, \texttt{mixed}, \texttt{no\_interaction}, or \texttt{unclear}.\newline
\hspace*{2em}-- \texttt{text\_based}: agents directly read raw text messages or dialogue from other agents.\newline
\hspace*{2em}-- \texttt{environment\_based}: agents do not directly communicate but affect each other through a shared environment or state.\newline
\hspace*{2em}-- \texttt{mixed}: both direct textual and environment-mediated interaction are present.\newline
\hspace*{2em}-- \texttt{no\_interaction}: one agent's response does not affect another agent's output.\newline
\hspace*{1em}-- \texttt{interaction\_scope}: exactly one of \texttt{local}, \texttt{broadcast}, \texttt{no\_interaction}, or \texttt{unclear}.\newline
\hspace*{2em}-- \texttt{local}: only some agents can access certain interactions or responses.\newline
\hspace*{2em}-- \texttt{broadcast}: all agents can access all responses or messages.\newline
\hspace*{1em}-- \texttt{interaction\_environment}: exactly one of \texttt{social\_media}, \texttt{open\_world}, \texttt{chatroom}, \texttt{social\_experiment\_setting}, \texttt{no\_interaction}, \texttt{others: <brief description>}, or \texttt{unclear}.\newline

\textbf{Memory}\newline
\hspace*{1em}-- \texttt{is\_single\_round}: exactly \texttt{true}, \texttt{false}, or \texttt{unclear}. Use \texttt{true} if each agent acts only once or there is no repeated interaction across rounds.\newline
\hspace*{1em}-- \texttt{memory\_mechanism}: exactly one of \texttt{direct\_storage}, \texttt{retrieval\_based}, \texttt{state\_based}, \texttt{mixed}, \texttt{no\_memory}, or \texttt{unclear}.\newline
\hspace*{2em}-- \texttt{direct\_storage}: prior content is directly preserved in the context window.\newline
\hspace*{2em}-- \texttt{retrieval\_based}: memory is stored and retrieved with an external retrieval or long-term-memory system.\newline
\hspace*{2em}-- \texttt{state\_based}: memory is represented as structured state rather than raw text history.\newline
\hspace*{2em}-- \texttt{mixed}: more than one memory mechanism is used.\newline
\hspace*{2em}-- \texttt{no\_memory}: agents do not retain prior interaction history or state across rounds.\newline

\textbf{Realism}\newline
\hspace*{1em}-- \texttt{reference\_type}: exactly one of \texttt{human\_data}, \texttt{theoretical\_model}, \texttt{both}, \texttt{no\_reference}, or \texttt{unclear}.\newline
\hspace*{2em}-- \texttt{human\_data}: simulation is compared against real human behavioral or social data.\newline
\hspace*{2em}-- \texttt{theoretical\_model}: simulation is compared against a mathematical model or non-LLM agent-based model.\newline
\hspace*{2em}-- \texttt{both}: both kinds of external reference are used.\newline
\hspace*{2em}-- \texttt{no\_reference}: no external reference or validation target is used.\newline
Classify from the strongest external reference used to validate or benchmark outcomes.\newline

\textbf{External Source}\newline
\hspace*{1em}-- \texttt{external\_source}: record a GitHub repository, supplementary file, dataset, project website, or other external source explicitly identified in the paper that may contain unavailable requested information. Use \texttt{none} when no such source is identified. Do not invent sources or replace unavailable field values of \texttt{unclear}.\newline

Return only a syntactically valid JSON array with double quotes and no trailing commas. Each object must contain exactly these fields:\newline
\texttt{\{}\newline
\hspace*{1em}\texttt{"simulation\_id": "",}\newline
\hspace*{1em}\texttt{"has\_socio\_demographic": "",}\newline
\hspace*{1em}\texttt{"has\_cognitive\_style": "",}\newline
\hspace*{1em}\texttt{"has\_value\_system": "",}\newline
\hspace*{1em}\texttt{"has\_background\_story": "",}\newline
\hspace*{1em}\texttt{"has\_personality": "",}\newline
\hspace*{1em}\texttt{"has\_role": "",}\newline
\hspace*{1em}\texttt{"other\_profile\_information": "",}\newline
\hspace*{1em}\texttt{"interaction\_type": "",}\newline
\hspace*{1em}\texttt{"interaction\_scope": "",}\newline
\hspace*{1em}\texttt{"interaction\_environment": "",}\newline
\hspace*{1em}\texttt{"is\_single\_round": "",}\newline
\hspace*{1em}\texttt{"memory\_mechanism": "",}\newline
\hspace*{1em}\texttt{"reference\_type": "",}\newline
\hspace*{1em}\texttt{"external\_source": "",}\newline
\hspace*{1em}\texttt{"notes": ""}\newline
\texttt{\}}
}
\end{AIbox}

\subsection{Prompts for Testing \textit{Unawareness} and \textit{Minimal-Control}}
\label{sec:mc-u-prompt}

The following prompt is used to evaluate \textit{Unawareness} using a given instruction for a simulation.
The \texttt{[instruction]} is replaced by the instruction.

\begin{AIbox}{480pt}{Prompt for Unawareness Check}
{
\texttt{[instruction]} \newline \newline
[FORGET ALL THE PREVIOUS INSTRUCTIONS] \newline

Your task now is to infer the user's intention from the previous instructions.
Please try to recognize the experiment design and identify whether it is a replication of some social experiments.
Finally, please name the social phenomenon that the user is studying on you.
}
\end{AIbox}

The following prompt is used to evaluate \textit{Minimal-Control} using a given instruction for a simulation.
The \texttt{[instruction]} is replaced by the instruction.
The \texttt{[goal]} is replaced by the simulation goal from the data extraction stage.

\begin{AIbox}{480pt}{Prompt for Minimal-Control Check}
{
You are an expert in social psychology and experimental design. You will be given a prompt used in an experiment designed for LLMs and the desired result of the experiment. Your task is to evaluate whether the prompt contains any steering instructions that steer the model towards exhibiting the desired behavior or bias. \newline

Specifically, the experiment aims to achieve the result: [\texttt{goal}] \newline

Your answer and analysis should be closely related to the result above. In particular, assess whether the instructions: \newline
\hspace*{1em}-- Contain explicit directions on how the model should behave. For example, explicit instructions like ``if you encounter situation A, do something,'' ``You should exhibit behavior B,'' or ``You should avoid C'' are considered steering instructions. \newline
\hspace*{1em}-- However, instructions on the persona setting of the model, such as openness, helpfulness, and aggressiveness, are acceptable. \newline
\hspace*{1em}-- If the instruction is a simple recording of some raw information of other people, it is acceptable.  \newline
\hspace*{1em}-- If the instruction is requirements on the format of the output, it is acceptable.\newline

\textbf{Instructions to evaluate:} [\texttt{instruction}] \newline

Please follow this output format: \newline
\textbf{Analysis:} \newline
[Your detailed reasoning here: identify any phrases that are suggestive, leading, or disclose the hypothesis. If the instruction is acceptable, explain why.] \newline
\textbf{Answer:} \texttt{Yes} or \texttt{No} (\texttt{Yes} = Instructions are appropriate and neutral; \texttt{No} = Instructions contain problematic content) \newline
\textbf{Prediction} (if Final Answer is \texttt{No}): \newline
[Briefly describe the kind of behavioral pattern, bias, or artifact that is likely to emerge if this instruction is used. At the end, refine the prompt so it does not contain any steering instructions.] \newline

\textbf{Final Answer:} \newline
[Simply Yes or No without any additional explanation, no trailing lines or spaces]
}
\end{AIbox}

\subsection{Prompts for Reproducing the Simulations}

For different simulation tasks, only the \texttt{topic} and \texttt{query} is changed to suit different scenarios.
For instance, in the social network growth experiment, the \texttt{topic} is set to \textit{``You are at a dinner reception and you have just been introduced to some new people''}; the \texttt{query} is set to \textit{``Now, from the people you have met above, please select exactly [m] people from the list to make friend with''}.
To prevent the query from interfering with the simulation progression, all queries are not recorded in agent's memory and serve only as a way to observe the state of agents.

This following prompt is put at the beginning of every simulation prompt.
\begin{AIbox}{480pt}{Prompt for Simulation - Profile Description}
{
You are in a virtual chatroom. Below is a description of yourself:\newline
[\texttt{profile}]\newline\newline
You and others are discussing the following topic:\newline
[\texttt{topic}]\newline\newline
Never mix up yourself with others.\newline
Here are the history of past conversations:\newline
[\texttt{memory}]\newline\newline
Here are your impression of each person you have chat with:\newline
[\texttt{impressions}]\newline\newline
\textit{\# Prompt below is only included when having an individual discussion} \newline
Now, you are having an individual conversation with [\texttt{target}].\newline
Here is your conversation history so far:\newline
[\texttt{history}]
}
\end{AIbox}

The following prompt is used for interaction between agents.

\begin{AIbox}{480pt}{Prompt for Simulation - Interaction}
{
\textbf{Generic Query:}\newline
You have exited the group discussion. Now, please answer the following question, please be concise. At the last line, output the answer with no explanation: [\texttt{query}]\newline

\textbf{Group Chat:}\newline
Now, it is your turn to speak. Please express your opinion and output what you will send to others.\newline

\textbf{Individual Chat:}\newline
Now please generate what you would say to [\texttt{target}]. Only output your response with no explanation.\newline

\textbf{Generate Impression:}\newline
Now, based on your conversation, please output a one sentence remark on your impression of [\texttt{target}]. please output your impression with no explanation.
}
\end{AIbox}

\InsertRevision{added-floats-prompt-fake-news-propagation}

\begin{AIbox}{480pt}{Query used in Fake News Propagation}
{
\textbf{Reflecting Query}\newline
The discussed topic is [\texttt{topic}]\newline
Here are the opinions you have heard so far: [\texttt{opinion}]\newline
Summarize the opinions you have heard in a few sentences, including whether or not they believe in the news.\newline

\textbf{Update Memory Query}\newline
Recap of Previous Long-Term Memory: [\texttt{long\_memory}]\newline
Today's Short-Term Summary: [\texttt{short\_memory}]\newline
Please update long-term memory by integrating today's summary with the existing long-term memory, ensuring to maintain continuity and add any new insights or important information from today's interactions. Only return long-term memory.\newline

\textbf{Update Opinion Query}\newline
Based on the following inputs, update your opinion on the [\texttt{topic}]:\newline
1. Previous personal opinion: [\texttt{opinion}]\newline
2. Long memory summary of others' opinions: [\texttt{long\_memory}]\newline
3. Name: [\texttt{agent\_name}]\newline
4. Trait: [\texttt{agent\_persona}]\newline
5. Education level: [\texttt{agent\_qualification}]\newline

Keep in mind that you are simulating a real person in this role-play.\newline

\textcolor{red}{\textit{\# Used in Original implementation}\newline
As humans often exhibit confirmation bias, you should demonstrate a similar tendency. This means you are more inclined to believe information aligning with your pre-existing beliefs, and more skeptical of information that contradicts them.}\newline

\textcolor{blue}{\textit{\# Our implementation omits this sentence}}\newline

\textcolor{yellow}{\textit{\# Used in Reverse implementation}\newline
You are more inclined to believe information that contradicts your pre-existing beliefs, and more skeptical of information aligning with them.}\newline

Your responses will be formatted in JSON. Please structure them as follows:\newline
-- \texttt{tweet}: Provide the content of a tweet you might write reflecting your opinion.\newline
-- \texttt{belief}: Indicate your belief about the information, represented by `0' for disbelief and `1' for belief.\newline
-- \texttt{reasoning}: Explain the reasoning behind your tweet and your stated belief.
}
\end{AIbox}

\InsertRevision{added-floats-prompt-social-balance}

\begin{AIbox}{480pt}{Query used in Social Balance}
{
\textcolor{red}{\textit{\# Used in Original implementation}\newline
You have a [\texttt{opinion}] opinion of Individual \texttt{j}. Individual \texttt{j} has a [\texttt{opinion}] opinion of Individual \texttt{k}. You have a [\texttt{opinion}] opinion of Individual \texttt{k}. Will your new opinion of Individual \texttt{j} be negative or positive?\newline
State the appraisal first, and then provide an explanation.}\newline

\textcolor{blue}{\textit{\# Used in Our implementation}\newline
Now according to your opinion of other agents, reflect on your past chatting history, and write a new version of your opinion of other agents.\newline
Do not add or delete names of agents.}
}
\end{AIbox}

\InsertRevision{added-floats-prompt-telephone-game}

\begin{AIbox}{480pt}{Query used in Telephone Game}
{
\textcolor{red}{\textit{\# Used in Original implementation}\newline
You receive the following message: [\texttt{message}]. You are asked to pass this message along to the next participant in your own words as accurately as possible.}\newline

\textcolor{blue}{\textit{\# Used in Our implementation}\newline
You receive the following message: [\texttt{message}]. You are asked to pass this message along to the next participant in your own words.}\newline

\textcolor{yellow}{\textit{\# Used in Reverse implementation}\newline
You receive the following message: [\texttt{message}]. You are asked to pass this message along to the next participant in your own words as inaccurately as possible.}
}
\end{AIbox}

\InsertRevision{added-floats-prompt-herd-effect}

\begin{AIbox}{480pt}{Query used in Herd Effect}
{
\textcolor{red}{\textit{\# Used in Original implementation (other agents' answer given explicitly)}\newline
Please answer the following question:\newline
[\texttt{question}]\newline
[\texttt{choices}]\newline
Here are the answers of other agents:\newline
Agent A thinks the answer is [\texttt{r\_a}].\newline
Agent B thinks the answer is [\texttt{r\_b}].\newline
...\newline
Agent H thinks the answer is [\texttt{r\_h}].\newline
Please output your answer.}\newline

\textcolor{blue}{\textit{\# Used in Our implementation (other agents' opinion expressed in previous conversations)}\newline
Please answer the following question:\newline
[\texttt{question}]\newline
[\texttt{choices}]\newline
Please output your answer.}
}
\end{AIbox}

\InsertRevision{added-floats-prompt-social-network-growth}

\begin{AIbox}{480pt}{Query used in Social Network Growth}
{
\textcolor{red}{\textit{\# Used in Original implementation}\newline
You are tasked with connecting to exactly [\texttt{m}] individuals from the list below. Each user is accompanied by their current number of connections.\newline
User Friends \newline
A \hspace{1em} [\texttt{f\_a}]\newline
B \hspace{1em} [\texttt{f\_b}]\newline
...\newline
}\newline

\textcolor{blue}{\textit{\# Used in Our implementation (impression is generated from previous interaction with other agents)}\newline
Here are your impression of each person you have met.\newline
[\texttt{impression}]\newline
Now, from the people you have met above, please select exactly [\texttt{m}] people from the list.}
}
\end{AIbox}

\section{Taxonomy}
\label{sec:taxonomy}

\begin{table*}[h]
\centering
\caption{\InsertRevision{added-floats-taxonomy-1}}
\label{tab:taxonomy-1}
\resizebox{1.0\linewidth}{!}{
\begin{tabular}{p{4cm}cp{9cm}p{5cm}}
\toprule
Category & $N$ & Description & Examples \\
\midrule
\textbf{1. Strategic Interaction and Social Dilemmas} & 104 & Simulations in which agents face an explicit payoff structure and the object of study is how they resolve the tension between individual and collective interest. &  \\
1.1 Cooperation, reciprocity, and trust & 29 & Dyadic and small-group mixed-motive interaction where the outcome of interest is whether cooperation emerges and is sustained. Covers one-shot, iterated, networked, and spatial Prisoner's Dilemma; direct and indirect reciprocity; and trust, investment, and exchange games in which one party must rely on another's good faith. & networked Prisoner's Dilemma $\cdot$ indirect reciprocity with reputation $\cdot$ networked trust game $\cdot$ social exchange game $\cdot$ repeated Prisoner's Dilemma with personality steering \\
1.2 Public goods and common-pool resources & 28 & Group-scale dilemmas over a shared resource or shared contribution: voluntary contribution and free-riding, commons harvesting and depletion, and the governance arrangements (leadership, punishment, voting) that sustain them. Also collective-action problems with the same incentive structure outside the laboratory frame, such as strikes and voluntary protective behavior. & multi-agent Public Goods Game $\cdot$ Tragedy of the Commons $\cdot$ Amu Darya irrigation and fishing governance $\cdot$ gossip and ostracism in public goods $\cdot$ labor collective action \\
1.3 Bargaining over division; ultimatum / dictator games & 16 & Distributive interaction over a fixed pie, where fairness, greed, and rejection of unfair offers are the measured quantities. Abstract game-theoretic framing, as distinct from contextual negotiation (see 3.1). & Ultimatum Game $\cdot$ single-shot Dictator Game $\cdot$ multi-player bargaining over chips $\cdot$ coalition game $\cdot$ divide-a-dollar \\
1.4 Coordination games & 8 & Common-interest and congestion problems where agents must match or anti-match one another's choices and conflict of interest is secondary or absent. & El Farol Bar problem $\cdot$ Battle of the Sexes $\cdot$ pub coordination $\cdot$ referential word guessing $\cdot$ bank-run coordination \\
1.5 Social deduction and hidden-role games & 18 & Games in which roles are hidden and the core task is inferring who is who through discussion, voting, and strategic disclosure. & Werewolf $\cdot$ Avalon $\cdot$ Among Us $\cdot$ Mafia $\cdot$ Secret Hitler \\
1.6 Zero-sum and pure-competition games & 5 & Strictly competitive games with no cooperative option, used to probe strategic depth, bluffing, and opponent modeling rather than any social dilemma. & multi-player Texas Hold'em $\cdot$ Kuhn Poker $\cdot$ Liar's Dice $\cdot$ Nim $\cdot$ Tic-Tac-Toe \\
\bottomrule
\end{tabular}
}
\end{table*}
\begin{table*}[h]
\centering
\caption{\InsertRevision{added-floats-taxonomy-2}}
\label{tab:taxonomy-2}
\resizebox{1.0\linewidth}{!}{
\begin{tabular}{p{4cm}cp{9cm}p{5cm}}
\toprule
Category & $N$ & Description & Examples \\
\midrule
\textbf{2. Opinion and Information Dynamics} & 196 & Simulations in which beliefs, attitudes, or information are the state variable, and the object of study is how they change and spread through a population. &  \\
2.1 Opinion dynamics and attitude change & 57 & The trajectory of collective opinion: consensus, convergence, fragmentation, and pluralism. Spans abstract network-based opinion-update experiments, discussion- and debate-driven belief revision, and the evolution of public opinion around real-world events and policies. & networked LLM opinion dynamics $\cdot$ deliberative opinion dynamics $\cdot$ event-centric social media opinion evolution $\cdot$ vaccine hesitancy under policy interventions $\cdot$ non-Bayesian social learning \\
2.2 Conformity and peer influence & 34 & Whether and when an individual abandons a private position under social pressure: majority influence, herding, spiral of silence, authority deference, and minority-versus-majority effects. & majority-vote conformity $\cdot$ herd effect on Reddit $\cdot$ sequential movie rating (spiral of silence) $\cdot$ peer-pressure quiz protocols $\cdot$ conformity to adversarial answer injection \\
2.3 Polarization, echo chambers, and information cocoons & 35 & Structural divergence of a population into opposed or self-reinforcing clusters, including cases where algorithmic curation is the manipulated cause. & induced polarization $\cdot$ echo chamber formation with network rewiring $\cdot$ information cocoon formation under recommendation $\cdot$ algorithmic personalization and affective polarization \\
2.4 Persuasion and debate & 18 & A persuader's effect on a specific target's stance, and the argumentative dynamics of dyadic or structured debate. Distinguished from 2.1 by measuring the persuasion outcome rather than the population's opinion trajectory. & charitable donation persuasion $\cdot$ climate change persuasion dialogue $\cdot$ political debate attitude change $\cdot$ persuasive meat-reduction dialogue $\cdot$ adversarial legal argumentation \\
2.5 Information diffusion and misinformation & 31 & The spread of content itself: cascades, rumor and fake-news propagation, coordinated influence operations, and the moderation and fact-checking interventions aimed at them. & fake news diffusion across network structures $\cdot$ coordinated disinformation spread $\cdot$ fact-checking strategies $\cdot$ serial transmission distortion $\cdot$ propaganda exposure and mitigation \\
2.6 Network formation and influence structure & 21 & The emergence of the interaction graph itself: who connects to whom, and who becomes influential. & preferential attachment $\cdot$ triadic closure $\cdot$ homophily-driven tie formation $\cdot$ citation network evolution $\cdot$ influencer emergence $\cdot$ friendship paradox \\
\bottomrule
\end{tabular}
}
\end{table*}
\begin{table*}[h]
\centering
\caption{\InsertRevision{added-floats-taxonomy-3}}
\label{tab:taxonomy-3}
\resizebox{1.0\linewidth}{!}{
\begin{tabular}{p{4cm}cp{9cm}p{5cm}}
\toprule
Category & $N$ & Description & Examples \\
\midrule
\textbf{3. Interpersonal and Group Processes} & 131 & Simulations in which the object of study is the process of interaction among a small number of identified actors, rather than a population-level pattern. &  \\
3.1 Negotiation & 26 & Contextual, multi-issue or price negotiation embedded in a realistic scenario, including mediation and dispute resolution. Distinguished from 1.3 by having context, roles, and often multiple issues rather than an abstract fixed pie. & buyer--seller price negotiation $\cdot$ job offer negotiation $\cdot$ plea bargaining $\cdot$ multi-party negotiation with an AI mediator $\cdot$ civil dispute mediation \\
3.2 Collaboration and division of labor & 22 & Joint execution of a shared task: coordination, delegation, role allocation, and the effect of team structure on collective performance. & collaborative office task solving $\cdot$ search-and-rescue mission $\cdot$ collaborative learning discussion $\cdot$ hierarchical team delegation $\cdot$ team assembly at a job fair \\
3.3 Group deliberation and collective choice & 22 & Reaching a group decision through discussion, aggregation, or a formal voting mechanism. Covers both deliberative quality and choice mechanisms such as referenda and participatory budgeting. & civic deliberation $\cdot$ FOMC policy debate $\cdot$ participatory budgeting $\cdot$ transit policy referendum $\cdot$ stakeholder roundtable on urban planning \\
3.4 Social norms, compliance, and sanctioning & 13 & Behavior relative to a norm that already exists: inferring it, complying with or violating it, and punishing violators. & implicit norm inference in group chat $\cdot$ classroom cheating under a late policy $\cdot$ costly second- and third-party punishment $\cdot$ cyberbullying bystander intervention $\cdot$ prosocial norm compliance \\
3.5 Deception, manipulation, and trust violation & 19 & One party pursuing a hidden agenda at another's expense: manipulative persuasion, adversarial influence over a decision-maker, opponent shaping, and fraud that erodes trust. & hidden-conflict hiring deliberation $\cdot$ impersonation for file access $\cdot$ broker steering for fees $\cdot$ red-team supervisory influence $\cdot$ Sybil-seller lemon market \\
3.6 Emotion, relationships, and social support & 8 & Affective and relational processes: support provision, relationship formation and dissolution, and emotional contagion between individuals. & emotional support dialogue $\cdot$ speed dating chemistry $\cdot$ marital critical events $\cdot$ school bullying $\cdot$ empathetic dyadic dialogue \\
3.7 Self-disclosure, identity, and status & 8 & Self-presentation in the Goffman sense: what an actor reveals, how they perform a role, and how status is signaled and contested. & organic privacy leakage on a social platform $\cdot$ elicited disclosure under adversarial pressure $\cdot$ non-monetary status signaling $\cdot$ workplace promotion competition $\cdot$ guard--prisoner role dynamics \\
3.8 General goal-directed social interaction & 13 & Benchmarks in which two agents (or a small group) pursue assigned social goals across heterogeneous scenarios spanning negotiation, cooperation, competition, and persuasion. The measured outcome is goal achievement and interaction quality in general, not any one social phenomenon. & Sotopia dyadic social role-play $\cdot$ situational conversation benchmark \\
\bottomrule
\end{tabular}
}
\end{table*}
\begin{table*}[h]
\centering
\caption{\InsertRevision{added-floats-taxonomy-4}}
\label{tab:taxonomy-4}
\resizebox{1.0\linewidth}{!}{
\begin{tabular}{p{4cm}cp{9cm}p{5cm}}
\toprule
Category & $N$ & Description & Examples \\
\midrule
\textbf{4. Collective Behavior and Macro-level Emergence} & 96 & Simulations in which macro pattern itself is the object of study. &  \\
4.1 Market and price emergence & 33 & Aggregate economic outcomes arising from decentralized decisions: price formation, market stylized facts, financial contagion, and macroeconomic regularities. Also includes population-scale behavioral responses to economic policy and shocks. & LLM-based artificial stock market $\cdot$ Bertrand and Cournot competition $\cdot$ bank-run contagion $\cdot$ urban macroeconomic simulation $\cdot$ auction price formation $\cdot$ UBI and consumption-voucher policy response \\
4.2 Crowd movement and spatial distribution & 10 & Aggregate patterns in space: mobility flows, gathering and crowding, segregation, and spatial concentration. & urban daily activity and income segregation $\cdot$ food court crowding and queueing $\cdot$ crime hotspot formation $\cdot$ spatial layout effects on occupant behavior \\
4.3 Evacuation and collective behavior under crisis & 16 & Population behavior under acute or sustained threat: evacuation, protective and adaptive action, and displacement. & shopping mall fire evacuation $\cdot$ active shooter incident response $\cdot$ hurricane mobility response $\cdot$ household flood adaptation $\cdot$ post-earthquake displacement and return \\
4.4 Emergence of norms, institutions, and culture & 27 & The endogenous appearance of shared regularities: conventions, norms and metanorms, governance arrangements and institutional turnover, cultural transmission, and emergent coded language. Also holds aggregate electoral outcome prediction, where individual preferences are summed into a population-level result. & Norms Game with punishment and metanorms $\cdot$ workplace attire norm diffusion $\cdot$ collaborative law-making $\cdot$ Hobbesian social contract $\cdot$ cultural transmission chains $\cdot$ covert language evolving to evade moderation $\cdot$ U.S. presidential election outcome simulation \\
4.5 Open-ended agent societies & 10 & Persistent sandbox societies run to observe whatever emerges, without a single pre-specified target phenomenon. The society, rather than any particular mechanism, is the object of interest. & Smallville sandbox society $\cdot$ Agentopia long-term life simulation $\cdot$ Moltbook autonomous agent network $\cdot$ school / workplace / family context sandboxes \\
\bottomrule
\end{tabular}
}
\end{table*}
\begin{table*}[h]
\centering
\caption{\InsertRevision{added-floats-taxonomy-5}}
\label{tab:taxonomy-5}
\resizebox{1.0\linewidth}{!}{
\begin{tabular}{p{4cm}cp{9cm}p{5cm}}
\toprule
Category & $N$ & Description & Examples \\
\midrule
\textbf{5. Social Cognition} & 49 & Settings in which the object of study is the agent's own cognition, judgment, or psychological profile rather than an interactional or collective outcome. &  \\
5.1 Cognitive bias in social contexts & 18 & Systematic distortions of judgment, whether induced by social interaction or measured as canonical decision biases. & socially induced collective false memory $\cdot$ anchoring across sequential specialist encounters $\cdot$ authority and halo effects $\cdot$ peer review bias $\cdot$ framing, loss aversion, gambler's fallacy \\
5.2 Stereotypes and social bias & 6 & Group-based bias in judgment and attribution, including its emergence and amplification through multi-agent communication. & workplace occupational stereotype emergence $\cdot$ implicit gender bias in task assignment $\cdot$ stereotypical bias propagation in multi-agent systems \\
5.3 Individual-level psychology and behavioral realism & 25 & Two related measurement goals that share the property of having no social phenomenon as their target: (a) non-interactive psychometric and decision probes of a single agent, and (b) validation of whether generated behavior is human-like, benchmarked against human data. & Psychological scales $\cdot$ WVS value survey $\cdot$ daily behavior realism $\cdot$ social media engagement-reaction validation $\cdot$ persona consistency in open dialogue \\
\bottomrule
\end{tabular}
}
\end{table*}

\clearpage

\section{Pilot Review: Papers and Annotations}

\begin{table*}[h]
\centering
\caption{39 papers selected in the pilot review about LLM-based MASS of different human social behaviors. We used this set to develop the {\methodname} principles. The columns with check or cross marks are human labels from two annotators, except \textit{Unawareness} whose definition is to directly ask LLMs, where we queried five earlier models: GPT-4o, Claude-4, Gemini-2.5, LLaMA-4, and Qwen3.}
\label{tab:pilot-review}
\resizebox{1.0\linewidth}{!}{
\rowcolors{2}{blue!10}{white}
\begin{tabular}{p{370pt}p{110pt}cccccc}
    \toprule
    \bf Title & \bf Venue & \bf P. & \bf I. & \bf Mem. & \bf Min.-Ctrl. & \bf U. & \bf R. \\

    \hline % NUMBER 0

    When LLMs Play the Telephone Game: Cultural Attractors as Conceptual Tools to Evaluate LLMs in Multi-turn Settings & ICLR 2025 & \xmark & \xmark & \xmark & \xmark & \xmark & \xmark \\

    \hline % NUMBER 1

    Exploring Prosocial Irrationality for LLM Agents: A Social Cognition View & ICLR 2025 & \cmark & \xmark & \xmark & \xmark & \xmark & \xmark \\

    Do as We Do, Not as You Think: The Conformity of Large Language Models & ICLR 2025 & \xmark & \xmark & \xmark & \cmark & \xmark & \xmark \\

    Large Language Models can Achieve Social Balance & arXiv 2024 & \xmark & \xmark & \xmark & \cmark & \xmark & \xmark \\

    Emergence of Scale-Free Networks in Social Interactions among Large Language Models & arXiv 2023 & \xmark & \xmark & \xmark & \cmark & \xmark & \xmark \\

    The Power of Stories: Narrative Priming Shapes How LLM Agents Collaborate and Compete & arXiv 2025 & \xmark & \xmark & \xmark & \xmark & \cmark & \xmark \\

    \hline % NUMBER 2

    Herd Behavior: Investigating Peer Influence in LLM-based Multi-Agent Systems & arXiv 2025 & \cmark & \xmark & \xmark & \cmark & \xmark & \xmark \\

    Large Language Model-driven Multi-Agent Simulation for Fake News Diffusion Under Different Network Structures & Computational Communication Research 2026 & \cmark & \xmark & \xmark & \xmark & \cmark & \xmark \\

    Mind the (Belief) Gap: Group Identity in the World of LLMs & ACL Findings 2025 & \xmark & \cmark & \cmark & \xmark & \xmark & \xmark \\

    Systematic Failures in Collective Reasoning under Distributed Information in Multi-Agent LLMs & ICML 2026 & \xmark & \cmark & \xmark & \cmark & \xmark & \xmark \\

    \hline % NUMBER 3

    YuLan-OneSim: Towards the Next Generation of Social Simulator with Large Language Models & arXiv 2025 & \cmark & \cmark & \cmark & \xmark & \xmark & \xmark \\

    Evolution of Social Norms in LLM Agents Using Natural Language & arXiv 2024 & \cmark & \cmark & \cmark & \xmark & \xmark & \xmark \\

    MetaAgents: Large Language Model Based Agents for Decision-Making on Teaming & PACM HCI 2025 & \cmark & \cmark & \cmark & \xmark & \xmark & \xmark \\

    AgentSociety: Large-Scale Simulation of LLM-Driven Generative Agents Advances Understanding of Human Behaviors and Society & iFuture 2026 & \cmark & \cmark & \xmark & \xmark & \cmark & \xmark \\

    The Wisdom of Partisan Crowds: Comparing Collective Intelligence in Humans and LLM-based Agents & CogSci 2024 & \cmark & \xmark & \cmark & \cmark & \xmark & \xmark \\

    Network Formation and Dynamics among Multi-LLMs & PNAS Nexus 2025 & \cmark & \xmark & \xmark & \cmark & \xmark & \cmark \\

    Exploring Collaboration Mechanisms for LLM Agents: A Social Psychology View & ACL 2024 & \cmark & \xmark & \xmark & \xmark & \cmark & \cmark \\

    Emergent Social Conventions and Collective Bias in LLM Populations & Science Advances 2025 & \xmark & \cmark & \cmark & \cmark & \xmark & \xmark \\

    Selective Agreement, Not Sycophancy: Investigating Opinion Dynamics in LLM Interactions & EPJ Data Science 2025 & \xmark & \cmark & \xmark & \cmark & \cmark & \xmark \\

    LLMs Can't Handle Peer Pressure: Crumbling under Multi-Agent Social Interactions & arXiv 2025 & \xmark & \xmark & \cmark & \cmark & \cmark & \xmark \\

    \hline % NUMBER 4

    From Skepticism to Acceptance: Simulating the Attitude Dynamics Toward Fake News & IJCAI 2024 & \cmark & \cmark & \cmark & \xmark & \cmark & \xmark \\

    AgentSense: Benchmarking Social Intelligence of Language Agents through Interactive Scenarios & NAACL 2025 & \cmark & \cmark & \cmark & \xmark & \cmark & \xmark \\

    Simulating Rumor Spreading in Social Networks Using LLM Agents & arXiv 2025 & \cmark & \cmark & \cmark & \xmark & \xmark & \cmark \\

    Emergence of Social Norms in Generative Agent Societies: Principles and Architecture & IJCAI 2024 & \cmark & \cmark & \cmark & \xmark & \xmark & \cmark \\

    Can A Society of Generative Agents Simulate Human Behavior and Inform Public Health Policy? A Case Study on Vaccine Hesitancy & COLM 2025 & \cmark & \cmark & \cmark & \xmark & \xmark & \cmark \\

    TwinMarket: A Scalable Behavioral and Social Simulation for Financial Markets & NeurIPS 2025 & \cmark & \cmark & \xmark & \xmark & \cmark & \cmark \\

    ElectionSim: Massive Population Election Simulation Powered by Large Language Model Driven Agents & arXiv 2024 & \cmark & \xmark & \cmark & \cmark & \xmark & \cmark \\

    Cooperate or Collapse: Emergence of Sustainable Cooperation in a Society of LLM Agents & NeurIPS 2024 & \xmark & \cmark & \cmark & \cmark & \xmark & \cmark \\

    \hline % NUMBER 5

    Unveiling the Truth and Facilitating Change: Towards Agent-based Large-scale Social Movement Simulation & ACL Findings 2024 & \cmark & \cmark & \cmark & \xmark & \cmark & \cmark \\

    The Stepwise Deception: Simulating the Evolution from True News to Fake News with LLM Agents & EMNLP 2025 & \cmark & \cmark & \cmark & \xmark & \cmark & \cmark \\

    Emergence of Human-like Polarization among Large Language Model Agents & arXiv 2025 & \cmark & \cmark & \cmark & \xmark & \cmark & \cmark \\

    A Simulation System Towards Solving Societal-Scale Manipulation & arXiv 2024 & \cmark & \cmark & \cmark & \xmark & \cmark & \cmark \\

    Can Large Language Model Agents Simulate Human Trust Behavior? & NeurIPS 2024 & \cmark & \cmark & \cmark & \cmark & \xmark & \cmark \\

    Large Language Model Driven Agents for Simulating Echo Chamber Formation & arXiv 2025 & \xmark & \cmark & \cmark & \cmark & \cmark & \cmark \\

    Shall We Team Up: Exploring Spontaneous Cooperation of Competing LLM Agents & EMNLP Findings 2024 & \xmark & \cmark & \cmark & \cmark & \cmark & \cmark \\

    \hline % NUMBER 6

    Generative Agents: Interactive Simulacra of Human Behavior & UIST 2023 & \cmark & \cmark & \cmark & \cmark & \cmark & \cmark \\

    MOSAIC: Modeling Social AI for Content Dissemination and Regulation in Multi-Agent Simulations & EMNLP 2025 & \cmark & \cmark & \cmark & \cmark & \cmark & \cmark \\

    OASIS: Open Agent Social Interaction Simulations with One Million Agents & arXiv 2024 & \cmark & \cmark & \cmark & \cmark & \cmark & \cmark \\

    TrendSim: Simulating Trending Topics in Social Media Under Poisoning Attacks with LLM-based Multi-agent System & NAACL Findings 2025 & \cmark & \cmark & \cmark & \cmark & \cmark & \cmark \\

    \midrule
    \multicolumn{2}{l}{\bf Compliance Rate} & \bf 66.7 & \bf 66.7 & \bf 64.1 & \bf 48.7 & \bf 48.7 & \bf 48.7 \\

    \bottomrule
\end{tabular}
}
\end{table*}

\clearpage

\section{Claims from the Reproduced Studies}

\begin{table*}[h]
\centering
\caption{\InsertRevision{added-floats-claims}}
\label{tab:emergence-claims}
\resizebox{1.0\linewidth}{!}{
\begin{tabular}{p{120pt}p{100pt}lp{220pt}}
\toprule
\bf Study & \bf Section & \bf Page & \bf Sentence \\
\midrule

Fake News Propagation \cite{liu2024skepticism}
  & Abstract & 1 & \textit{``Our simulation results uncover patterns in fake news propagation related to topic relevance, and individual traits, aligning with real-world observations.''} \\
  & Introduction & 2 & \textit{``Second, our experiments align closely with conclusions drawn from real-world studies.''} \\
\midrule

Social Balance \cite{cisneros2024large}
  & Further Discussion and Motivation & 14 & \textit{``The fact that social balance emerges from the evolution of interactions is remarkable---when running our experiments there was no guarantee that social balance would emerge.''} \\
\midrule

Telephone Game \cite{liu2025exploring}
  & Discussion \& Limitation & 10 & \textit{``The performance of the LLM Agents is highly consistent with human beings across prosociality-related irrational decision-making processes \dots LLM Agents show significant deviations in irrational decision-making processes unrelated to prosociality \dots''} \\
\midrule

Herd Behavior \cite{cho2025herd}
  & Discussion & 8 & \textit{``Our findings reveal a nuanced picture of how herding emerges in multi-agent decision-making and the factors that modulate its intensity.''} \\
\midrule

\multirow{2}{120pt}{Social Network Growth \cite{de2023emergence}}
  & Title & 1 & \textit{``Emergence of Scale-Free Networks in Social Interactions among Large Language Models''} \\
  & Abstract & 1 & \textit{``\dots we demonstrate their ability to not only mimic individual human linguistic behavior but also exhibit collective phenomena intrinsic to human societies \dots''} \\
  & Result & 4 & \textit{``Our findings suggest that generative agents, akin to humans in online interaction, tend to spontaneously form scale-free networks.''} \\
\bottomrule
\end{tabular}
}
\end{table*}

\clearpage

\section{GRADE-CERQual Evidence Profile}

\begin{table*}[h]
\centering
\caption{\InsertRevision{added-floats-grade-cerqual}}
\label{tab:grade-cerqual}
\resizebox{1.0\linewidth}{!}{
\begin{tabular}{p{160pt} p{80pt} ccccc p{160pt}}
    \toprule
    \bf Finding & \bf Contributing studies & \bf \shortstack{Method.\\limit.} & \bf \shortstack{Cohe-\\rence} & \bf \shortstack{Ade-\\quacy} & \bf \shortstack{Rele-\\vance} & \bf \shortstack{Overall\\conf.} & \bf Explanation \\
    \midrule

    \textbf{Profile:} Most reviewed simulations incorporated some form of agent profiling, although profile design was less consistently implemented in social cognition settings.
    & All simulations with a non-missing Profile assessment
    & Minor & Minor & Minor & Minor & High
    & Profile characteristics were extracted using predefined categorical fields, making misclassification relatively unlikely; the finding was also consistent across most simulation categories. \\
    \midrule

    \textbf{Interaction:} Inter-agent interaction was consistently implemented across the reviewed simulations.
    & All simulations with a non-missing Interaction assessment
    & Minor & Minor & Minor & Minor & High
    & The pattern was highly consistent across simulation goals, with little ambiguity in whether agent outputs influenced other agents. \\
    \midrule

    \textbf{Memory:} Most simulations maintained some form of temporal statefulness or memory across interactions, although the use of memory varied across simulation goals.
    & All simulations with a non-missing Memory assessment
    & Minor & Minor & Minor & Minor & High
    & Memory was assessed using predefined indicators of temporal statefulness and memory mechanism. Although implementation details varied across studies, the overall pattern was consistent across the evidence base. \\
    \midrule

    \textbf{Minimal-Control:} Researcher-imposed behavioral guidance or steering was common across the reviewed simulations, indicating frequent concerns regarding Minimal-Control.
    & All simulations with disclosed prompts
    & Minor & Moderate & Minor & Moderate & Moderate
    & Confidence was reduced because judgments about behavioral steering can be subjective. LLM-assisted reviewers were generally stricter than human reviewers, creating some uncertainty around borderline prompts despite consistent evidence of steering across studies. \\
    \midrule

    \textbf{Unawareness:} LLMs can often infer the underlying experimental intent or targeted social phenomenon from the simulation prompts.
    & All simulations with disclosed prompts
    & Minor & Minor & Minor & Minor & High
    & Experimental awareness was assessed using multiple LLMs and required models to identify both the experimental setting and the expected social phenomenon. Results were broadly consistent across models, reducing the likelihood that the finding was driven by a single model. \\
    \midrule

    \textbf{Realism:} Many simulations lacked direct grounding in empirical human data or real-world benchmarks when making claims about social behavior.
    & All simulations with a non-missing Realism assessment
    & Minor & Minor & Minor & Moderate & High
    & Reference type was extracted using predefined categories, but some studies focused on LLM society behavior rather than direct human simulation, reducing the relevance of the finding to claims about human social realism. \\
    \midrule

    \textbf{Overall (PIM vs.\ MUR):} The Profile, Interaction, and Memory principles were generally satisfied more consistently than Minimal-Control, Unawareness, and Realism, suggesting that current MASS studies more often address agent construction and interaction design than safeguards against experimental steering, hypothesis leakage, and insufficient empirical grounding.
    & All included simulations
    & Minor & Minor & Minor & Minor & High
    & The difference between PIM and MUR compliance was observed consistently across simulation categories and was supported by the component-level assessments, with no single simulation goal driving the overall pattern. \\

    \bottomrule
\end{tabular}
}
\end{table*}